%% file: main.tex
\documentclass{article}

\usepackage{microtype}
\usepackage{graphicx}
\usepackage{subcaption}
\usepackage{booktabs} 

\usepackage{hyperref}

\usepackage[accepted]{icml2026}

\usepackage{amsmath}
\usepackage{amssymb}
\usepackage{mathtools}
\usepackage{amsthm}

\usepackage[utf8]{inputenc} 
\usepackage[T1]{fontenc}    
\usepackage{hyperref}       
\usepackage{url}            
\usepackage{booktabs}       
\usepackage{amsfonts}       
\usepackage{nicefrac}       
\usepackage{microtype}      
\usepackage[dvipsnames]{xcolor}         
\usepackage{booktabs}
\usepackage{tabularx}
\usepackage{multirow}
\usepackage{bbding}
\usepackage{tikz}
\usepackage{pifont}
\usepackage{algorithm}
\usepackage{algorithmic}
\usepackage{wrapfig}
\usepackage{cancel}
\usepackage{arydshln}

\usepackage[capitalize,noabbrev]{cleveref}

\theoremstyle{plain}
\newtheorem{theorem}{Theorem}[section]

\theoremstyle{definition}

\theoremstyle{remark}

\usepackage[textsize=tiny]{todonotes}

\begin{document}

\twocolumn[
  \icmltitle{Smoothie: Smoothing Diffusion on Token Embeddings for Text Generation}



  \icmlsetsymbol{equal}{*}

  \begin{icmlauthorlist}
    \icmlauthor{Alexander Shabalin}{cub,hse}
    \icmlauthor{Viacheslav Meshchaninov}{cub}
    \icmlauthor{Dmitry Vetrov}{cub}
  \end{icmlauthorlist}

  \icmlaffiliation{cub}{Constructor University, Bremen, Germany}
  \icmlaffiliation{hse}{HSE University, Moscow, Russia}

  \icmlcorrespondingauthor{Alexander Shabalin}{amshabalin@hse.ru}

  \icmlkeywords{Diffusion models, Text diffusion models, Continuous text diffusion models}

  \vskip 0.3in
]



\printAffiliationsAndNotice{}  

\begin{abstract}
  Diffusion models have achieved state-of-the-art performance in generating images, audio, and video, but their adaptation to text remains challenging due to its discrete nature. Prior approaches either apply Gaussian diffusion in continuous latent spaces, which inherits semantic structure but struggles with token decoding, or operate in categorical simplex space, which respect discreteness but disregard semantic relation between tokens. In this paper, we propose \underline{Smooth}ing D\underline{i}ffusion on Token \underline{E}mbeddings (\textsc{Smoothie}), a novel diffusion method that combines the strengths of both approaches by progressively smoothing token embeddings based on semantic similarity. This technique enables gradual information removal while maintaining a natural decoding process. Experimental results on several sequence-to-sequence and unconditional generation tasks demonstrate that \textsc{Smoothie} outperforms existing diffusion-based models in generation quality. Furthermore, ablation studies show that our proposed diffusion space yields better performance than both the standard embedding space and the categorical simplex. The code is available at \href{https://github.com/ashaba1in/smoothie}{\texttt{https: //github.com/ashaba1in/smoothie}}.
\end{abstract}

\input{sources/introduction}

\input{sources/background}

\input{sources/related_work}

\input{sources/method}

\input{sources/experiments}

\input{sources/limitations}

\input{sources/conclusion}

\input{sources/acknowledgments}

\input{sources/societal_impact}

\bibliography{bibliography}
\bibliographystyle{icml2026}

\newpage

\appendix
\onecolumn
\input{sources/appendix}

\end{document}

%% file: sources/introduction.tex
\section{Introduction}

Diffusion models attracted a lot of attention in recent years as they show very high generation quality in image \citep{latentdiffusion, sdxl}, audio \citep{stable_audio} and video \citep{stable_video_diffusion} domains surpassing all previous approaches such as GANs \citep{gan} and Normalizing Flows \citep{norm_flow}. Diffusion models work by introducing a forward process that gradually degrades an object by injecting Gaussian noise into it, and then learning the reverse process by denoising the object.

Applying diffusion models to text is challenging due to its discrete nature. Nevertheless, several works have explored ways to design suitable diffusion processes. One line of research proposes gradually removing information by replacing tokens with others sampled from a categorical distribution \citep{d3pm, diffusionbert, sedd}. Another approach applies Gaussian diffusion to the latent space of token embeddings \citep{diffusion-lm, diffuseq}. Additionally, some studies leverage the discreteness of text by performing diffusion directly on the vocabulary probability simplex instead of the embedding space \citep{han2022ssd, tess}.

Each of the described methods offers distinct advantages and limitations, as summarized in Table~\ref{tab:high_level_comparison}. Gaussian diffusion progressively removes semantic information: under the Euclidean semantic space hypothesis \citep{euclidean_space_hyp}, the distinguishability of noisy tokens depends on their initial distances in the latent space. The addition of Gaussian noise gradually disrupts these distances, making the semantics of a latent representation increasingly difficult to recover.
However, Gaussian diffusion does not account for the discrete nature of text, which complicates the mapping of generated latent vectors back to discrete tokens \citep{diffusion-lm, tencdm}.

On the other hand, categorical and simplex-based diffusion methods naturally preserve the discreteness of text and eliminate the need for an explicit decoding step. Nevertheless, they disregard semantic relationships between tokens during the noising process, resulting in a more erratic and less meaningful degradation of information.

\begin{figure*}[t]
  \begin{center}
    \includegraphics[width=0.9\linewidth]{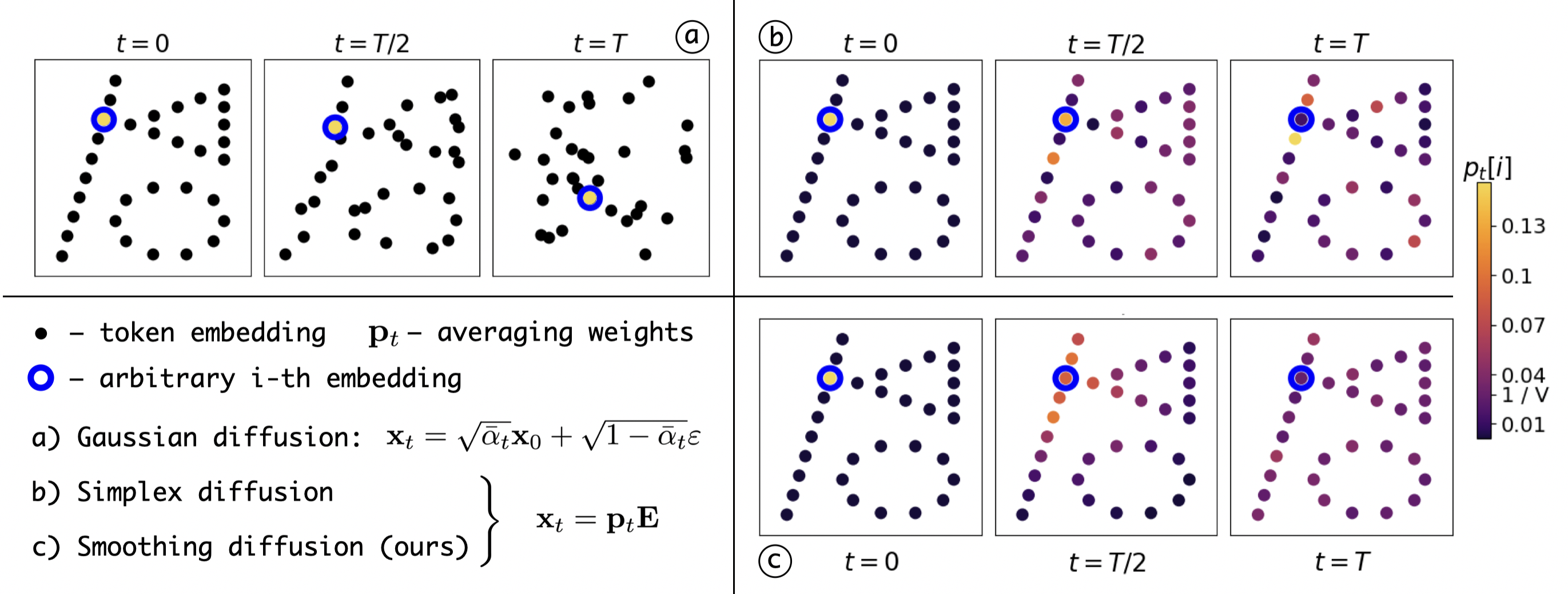}
  \end{center}
  \caption{An illustration of the diffusion process for Gaussian, simplex, and smoothing diffusion methods. The key distinction between simplex and smoothing diffusion is that the latter incorporates semantic relationships between tokens during the noise addition process.}
  \label{fig:toy_example}
\end{figure*}

\begin{table*}[t]
\centering
\caption{Comparison of diffusion methods in terms of accounting for text discreteness and semantics.}
\label{tab:high_level_comparison}
\begin{tabular}{lcccc}
\toprule
 & Categorical & Gaussian & Simplex & Smoothing (ours)\\
\midrule
Accounting for Discreteness & \color{Green}{\ding{51}} & \color{Red}{\ding{55}} & \color{Green}{\ding{51}} & \color{Green}{\ding{51}}\\
Accounting for Semantics & \color{Red}{\ding{55}} & \color{Green}{\ding{51}} & \color{Red}{\ding{55}} & \color{Green}{\ding{51}}\\
\bottomrule
\end{tabular}
\end{table*}

In this paper, we propose \textsc{Smoothie}, a smoothing diffusion framework that satisfies both properties. We represent each token with a vector based on distances between token embeddings. During the forward process, our diffusion mechanism gradually perturbs these distances, progressively dissolving semantic information. Like simplex diffusion, our method enables natural decoding from latent representations back to tokens. In theory, \textsc{Smoothie} is applicable not only to text, but to any domain where data comes from a categorical distribution with inherent similarity between categories (e.g. graphs or protein sequences).

We evaluate \textsc{Smoothie} on one unconditional and four sequence-to-sequence generation tasks and show that it outperforms existing diffusion-based approaches. Ablation studies further demonstrate that our method enables effective control over the trade-off between fluency and diversity of the generated text.

The main contributions of our work are as follows:
\begin{itemize}
    \item We propose a novel text diffusion framework that simultaneously respects the discrete nature of text and progressively removes semantic information from token representations during the forward process.
    \item We demonstrate the superior empirical performance of our approach other previous methods across multiple text generation tasks.
    \item We propose using a sampling method that, unlike the DDPM solver \citep{ddpm}, ignores the previous hidden state when computing the updated hidden state. This method has been shown to consistently outperform DDPM on all sequence-to-sequence tasks used for evaluation.
\end{itemize}

%% file: sources/background.tex
\section{Problem Statement and Background}

\paragraph{Problem Statement} In this work, we develop a model for both unconditional and sequence-to-sequence generation tasks. In all cases, the objective is to generate a target sequence $\mathbf{w}^y = {w^y_1, \dots, w^y_m}$. For sequence-to-sequence generation, the model additionally conditions on a source sequence $\mathbf{w}^x = {w^x_1, \dots, w^x_n}$. We assume access to parallel datasets, where each source sequence is paired with its corresponding target sequence.

\paragraph{Gaussian Diffusion Model} The diffusion process is defined in terms of a forward (noising) and a reverse (denoising) processes. Given an initial data point sampled from the data distribution, $\mathbf{x}_0 \sim p_{\text{data}}$, the forward process generates a sequence of progressively noisier latent variables $\mathbf{x}_1, \dots, \mathbf{x}_T$. Each step in this sequence is defined by the transition $\mathbf{x}_t \sim q(\mathbf{x}_t \mid \mathbf{x}_{t-1}) = \mathcal{N}(\sqrt{\alpha_t} \mathbf{x}_{t-1}, \sqrt{1 - \alpha_t},\varepsilon)$, where the parameter $\alpha_t \in [0, 1)$ controls the amount of noise injected at timestep $t$. This formulation also supports a direct sampling of $\mathbf{x}_t$ from $\mathbf{x}_0$ using the marginal distribution $q(\mathbf{x}_t \mid \mathbf{x}_0) = \mathcal{N}(\sqrt{\bar{\alpha}_t} \mathbf{x}_0, \sqrt{1 - \bar{\alpha}_t},\varepsilon)$, where $\bar{\alpha}_t = \prod_{s=0}^t \alpha_s$ is the cumulative product of noise scales. 

After the forward process is complete, a neural network $f_{\theta}$ is trained to reverse it by predicting the original data point $\mathbf{x}_0$ from the noisy input $\mathbf{x}_t$. During generation, the model iteratively denoises an initial sample $\mathbf{x}_T \sim \mathcal{N}(0, I)$, gradually reconstructing the data through the learned reverse process until it recovers $\mathbf{x}_0$.

\paragraph{Embedding Diffusion} The most popular continuous text diffusion approaches create a latent space by mapping tokens to their embeddings \citep{diffusion-lm, seqdiffuseq, diffuseq}. Then the Gaussian diffusion process is used to corrupt a latent. The decoding is usually performed by mapping a generated embedding to the token corresponding to the closest embedding.

\paragraph{Simplex Diffusion} SSD-LM \citep{han2022ssd} and TESS \citep{tess} propose a simplex diffusion model. They map each token $w$ to a $k$-logit simplex $\mathbf{s}^w \in \{\pm k\}^{V}$, where $V$ is the size of the vocabulary and 
\begin{align}\label{eq:simplex}
    \mathbf{s}_{(i)}^w = \begin{cases}
        + k, &i = w\\
        -k, &\text{otherwise}
    \end{cases}
\end{align}
Then the latent is represented as a sequence $\mathbf{S}_0 = (\mathbf{s}^{w^y_1}, \dots, \mathbf{s}^{w^y_m})$. Corruption is performed with the Gaussian diffusion process with noise variance multiplied by $k^2$ ($k=5$ by default), $\mathbf{S}_t = \sqrt{\bar{\alpha}_t} \mathbf{S}_0 + k\sqrt{1 - \bar{\alpha}_t} \varepsilon$. The model input is calculated by first producing a probability simplex over vocabulary, $\mathbf{p}_t = \operatorname{softmax}(\mathbf{S}_t)$, and then averaging token embeddings with obtained weights, $\mathbf{p}_t \mathbf{E}$, where $\mathbf{E}$ is a matrix of token embeddings.

%% file: sources/related_work.tex
\section{Related work}

Since the initial attempt to apply diffusion models to text generation \citep{multinomial_diffusion}, numerous studies have explored ways to better align the diffusion process with the specifics of textual data. D3PM \citep{d3pm} tried exploiting the semantic property of tokens by applying a discrete diffusion process that replaces tokens with semantically similar alternatives with higher probability. However, their experiments showed that simple token masking approach produces better empirical results.

Diffusion-LM \citep{diffusion-lm} proposed applying Gaussian diffusion in the continuous latent space of token embeddings, while TEncDM \citep{tencdm} further demonstrated that context-dependent embeddings provide a more suitable latent space for diffusion. Despite achieving strong generation quality, the downside of these methods is the requirement of an additional latent decoding step.

DiffuSeq-v2 \citep{diffuseq2} attempted to bridge the gap between discrete and continuous diffusion models by combining masking with Gaussian noise during the noising process. Another research direction \citep{han2022ssd, tess} focuses on mapping tokens to almost-one-hot simplex representations over the vocabulary and introducing Gaussian noise directly into this space. While this approach does not account for token semantics during noising, it preserves the discrete structure of text.

Our work is inspired by a different line of research developed in the image domain \citep{ihdm, blurring_diffusion}, where semantic information is gradually removed by smoothing pixel values according to the heat dissipation principle. This approach, however, cannot be directly applied to text due to its inherently discrete nature.

%% file: sources/method.tex
\section{Smoothing Diffusion}\label{sec:method}

In this section, we introduce \textsc{Smoothie}, a smoothing text diffusion model that incorporates both the discrete nature of text and the semantic relationships between tokens into the diffusion process. We will first derive the diffusion process for unconditional generation and then extend it to conditional generation. We provide an intuitive illustration of our approach, along with pseudo-code for the training and sampling procedures, in Fig.~\ref{fig:toy_example}, Alg.~\ref{alg:training}, and Alg.~\ref{alg:sampling}, respectively.

\subsection{Forward Diffusion Process}

Let $V$ denote the vocabulary size, and let $\mathbf{E} \in \mathbb{R}^{V \times d}$ be a fixed embedding matrix, where each row corresponds to a $d$-dimensional token embedding. To construct a latent space suitable for diffusion, we represent each token $w_i^y$ in a target sequence $\mathbf{w}^y$ with a vector of negative squared Euclidean distances between an embedding of token $w_i^y$ and embeddings of all tokens in the vocabulary:
\begin{align}\label{eq:D_0}
\mathbf{D}_0 = \mathbf{D}_0(\mathbf{E}_{\mathbf{w}^y}) = \left\{-\frac{\|\mathbf{E}_{w^y_i} - \mathbf{E}_j\|^2}{2} \right\}_{i,j=1}^{m,V}
\end{align}

Here, $m$ is the sequence length, $\mathbf{E}_{w^y_i}$ is the embedding of the i-th token in the sequence, and $\mathbf{E}_{j}$ is the embedding of the j-th vocabulary token. To generate a trajectory of progressively noisier latents, we define a non-Markovian forward, or noising process:
\begin{align}\label{eq:forward}
    q(\mathbf{D}_{1:T} | \mathbf{D}_{0}) = \prod_{t=1}^T q(\mathbf{D}_t | \mathbf{D}_0) = \prod_{t=1}^T \mathcal{N}\left(\mathbf{D}_t \bigg| \frac{1}{\sigma_t^2}\mathbf{D}_0, \delta^2 I\right)
\end{align}
The noise scheduler $\sigma_t$ ($1 < \sigma_1 < \dots < \sigma_T$) controls the amount of noise added at each timestep. The hyperparameter $\delta$ controls the stochasticity of the diffusion process and makes it non-deterministic. Following \citet{ihdm}, we keep $\delta$ independent of the timestep $t$.

To construct the model input, we convert $\mathbf{D}_t$ into a probability distribution over the vocabulary using the softmax function: $\mathbf{p}_t = \operatorname{softmax}(\mathbf{D}_t)$. In this formulation, each token is represented by the weights of Nadaraya-Watson kernel estimator applied over all embeddings in the vocabulary with Gaussian kernel whose bandwidth is defined by $\sigma_t$. The choice of a Gaussian kernel is motivated by the Euclidean semantic space hypothesis \citep{euclidean_space_hyp}, which assumes that semantic similarity correlates with Euclidean proximity in embedding space. As a result, as $\sigma_t$ increases, the probability mass initially centered in a single token gradually distributes between all other tokens, starting with the most semantically similar and ending with the most distant ones (see Fig. \ref{fig:toy_example} (c)). We show an example of this behavior in Appendix \ref{sec:semantic_degradation}.

Note that our approach can be viewed as a generalization of a simplex-based diffusion \citep{han2022ssd, tess}. In particular, by replacing our Euclidean distance with trivial metric, we get the latent space formulation defined in Eq. \ref{eq:simplex}, which ignores the semantic relationships between tokens. We prove this statement in Appendix \ref{sec:smoothie_tess}. In Section \ref{sec:experiments}, we show that incorporating semantic similarity into the diffusion process is crucial for achieving better performance.

\subsection{Reverse Diffusion Process}

The reverse, or denoising process, starts with a sample from prior distribution $p(\mathbf{D}_T)$ and ends with the denoised data sample $\mathbf{D}_0$. We define it as a Markov chain with Gaussian distributions:
\begin{align}\label{eq:reverse}
    p_{\theta}(\mathbf{D}_{0:T}) &= p(\mathbf{D}_T) \prod_{t=1}^T p_{\theta}(\mathbf{D}_{t-1} | \mathbf{D}_t)\\
    &= p(\mathbf{D}_T) \prod_{t=1}^T \mathcal{N}\big(\mathbf{D}_{t-1} | \mu_{\theta}(\mathbf{p}_t, t), \tilde{\delta}^2 I\big), &
\end{align}
where $\theta$ are trainable model parameters and $\tilde{\delta}^2$ is a noise variance used in the reverse process. Inspired by \citet{ihdm}, we allow noise variance to change between the forward and reverse processes. That permits us to explicitly control the stochasticity of the generation trajectory, which significantly affects the model performance. We discuss this effect in Section \ref{sec:delta}.

Our goal is to find such parameters $\theta$, that minimize the marginal negative likelihood of data samples $p_\theta(\mathbf{D}_{0}) = \int p_\theta(\mathbf{D}_{0:T}) \mathrm{d} \mathbf{D}_{1:T}$. We optimize the negative log-likelihood by minimizing its variational upper bound, which results in the following loss function\footnote{For complete derivation of the loss function see Appendix \ref{sec:loss_function}.}:
\begin{align}
L(\theta) = \mathbb{E}_q \left[\frac{1}{2\tilde{\delta}^2} \left\|\frac{1}{\sigma_t^2} \mathbf{D}_0 - \mu_{\theta}(\mathbf{p}_t, t)\right\|^2\right]
\end{align}
 
This implies that the most direct parameterization of $\mu_{\theta}$ is a model that predicts $\mathbf{D}_0 / \sigma_t^2$, corresponding to the posterior mean of the forward process. For practical reasons, we instead parameterize $\mu_{\theta}$ as $g_{\theta} / \sigma_t^2$ which ensures that all model outputs are scaled to have the same variance across timesteps. Also, following \citet{ddpm}, we simplify the objective by removing the scaling coefficient $2\tilde{\delta}^2\sigma_t^4$.
\begin{align}\label{eq:D_0_loss}
L_{\mathbf{D}}(\theta) = \mathbb{E}_{\mathbf{w}^y, t, \mathbf{p}_t} \left[\|\mathbf{D}_0(\mathbf{E}_{\mathbf{w}^y}) - g_{\theta}(\mathbf{p}_t, t)\|^2 \right]
\end{align}
However, this loss function is challenging to optimize due to the high variance and dimensionality of $\mathbf{D}_0$. To address this issue, we introduce the following theorem:

\begin{theorem}\label{th:theorem}
Let $g^*(\mathbf{p}_t, t)$ be an optimal prediction for Eq. \ref{eq:D_0_loss}. Then $g^*(\mathbf{p}_t, t) = \mathbf{D}_0(f^*(\mathbf{p}_t, t)) + C$, where $C$ is a constant that does not depend on $f^*(\mathbf{p}_t, t)$ and $f^*(\mathbf{p}_t, t)$ is an optimal prediction for Eq. \ref{eq:e_loss}
\begin{align}\label{eq:e_loss}
L_{\mathbf{E}}(\theta) = \mathbb{E}_{\mathbf{w}^y, t, \mathbf{p}_t} \left[\|\mathbf{E}_{\mathbf{w}^y} - f_{\theta}(\mathbf{p}_t, t)\|^2 \right]
\end{align}
\end{theorem}

We train the model $f_\theta$ by minimizing Eq.~\ref{eq:e_loss}. During the sampling, we start from the random noise $\mathbf{D}_T \sim \mathcal{N}(0, \tilde{\delta}^2 I)$ and iteratively update it using the following scheme to get a clean sample:
\begin{align}\label{eq:sampling}
\mathbf{D}_{t-1} = \frac{1}{\sigma_{t-1}^2} \mathbf{D}_0(f_{\theta}(\mathbf{p}_t, t)) + \tilde{\delta} \varepsilon, 
\end{align}
We emphasize that although the correct sampling scheme suggests using $g_{\theta}(\mathbf{p}_t, t)$ instead of $\mathbf{D}_0(f_{\theta}(\mathbf{p}_t, t))$, by the Theorem~\ref{th:theorem} both options are equivalent. This is because the resulting updates of these two schemes differ by a constant shift and the model on each step takes $\mathbf{p}_t = \operatorname{softmax}(\mathbf{D}_t)$ as input, which is invariant to shifts of $\mathbf{D}_t$. The proof of Theorem~\ref{th:theorem} is provided in Appendix \ref{sec:proof_4.1}.


Related methods such as SSD-LM \citep{han2022ssd} and TESS \citep{tess} employ cross-entropy loss during training. While our method is also compatible with this loss, in our experiments it led to inferior performance and faster overfitting. Therefore, we chose to rely on the MSE objective.

\begin{table}
\vspace{-5pt}
\begin{minipage}[t]{.48\textwidth}
\begin{algorithm}[H]
   \caption{Training}
   \label{alg:training}
\begin{algorithmic}
\STATE {\bfseries Input:} $\delta, \mathbf{w}^x, \mathbf{w}^y, t \sim \mathcal{U}(1, T), \varepsilon \sim \mathcal{N}(0, I)$
\STATE Compute $\mathbf{D}_0$ with Eq. \ref{eq:D_0}
\STATE Compute $\mathbf{D}_t = \mathbf{D}_t / \sigma_t^2 + \delta \varepsilon$
\STATE Compute $\mathbf{p}_t = \operatorname{softmax}(\mathbf{D}_t)$
\STATE Minimize $\| \mathbf{E}_{\mathbf{w}^y} - f_\theta(\mathbf{p}_t, t, \mathbf{w}^x) \|^2$
\end{algorithmic}
\end{algorithm}
\end{minipage}
\hfill
\begin{minipage}[t]{.48\textwidth}
\begin{algorithm}[H]
   \caption{Sampling}
   \label{alg:sampling}
\begin{algorithmic}
\STATE {\bfseries Input:} Source text $\mathbf{w}^x$, model $f_{\theta}$, noise std $\tilde{\delta}$
\STATE Sample $\mathbf{D}_T \sim \mathcal{N}(0, \tilde{\delta}^2 I)$
\FOR{$t$ in $\{T, \dots, 1\}$}
\STATE Compute $\mathbf{p}_t = \operatorname{softmax}(\mathbf{D}_t)$
\STATE Compute $\mathbf{D}_{t-1}$ with Eq. \ref{eq:sampling}
\ENDFOR \\
Decode tokens $\mathbf{\hat{w}}^y = \operatorname{argmax}(\mathbf{D}_0)$
\end{algorithmic}
\end{algorithm}
\end{minipage}
\end{table}

\subsection{Noise Scheduler}
The noise scheduler plays a crucial role in the diffusion process by controlling the rate at which the signal decays over time. Following the observation that text diffusion models benefit from adding more noise at the early stages of the forward process~\citep{tencdm}, we define our noise schedule as follows:
\begin{align}
\sigma_t = \left(\sigma_{\text{max}} - \sigma_{\text{min}}\right)\frac{2}{\pi}\arctan\left(\frac{1}{d} \sqrt{\frac{t}{T - t + \epsilon}}\right) + \sigma_{\text{min}}
\end{align}

Here, $t \in [0, T]$, $\sigma_{\text{min}}$ and $\sigma_{\text{max}}$ set the minumum and maximum bandwidth respectively, $d$ controls the rate of noise accumulation, and $\epsilon$ is a small constant added to prevent division by zero. Throughout our experiments, we use $\sigma_{\text{min}} = 1.5$, $\sigma_{\text{max}} = 200$ and $d \in \{5, 7\}$ to achieve a linear increase in model entropy with increasing $t$ \citep{cdcd}. Also, we set $\delta = 1$ during training. We discuss the noise scheduler ablation in Appendix \ref{sec:noise_scheduler}.

\subsection{Self-conditioning}
Following previous works \citep{cdcd, seqdiffuseq, tencdm}, we employ \textit{self-conditioning} \citep{self-cond} to our model. During training, with 50\% probability the model is fed with self-condition set to zero: $\mathbf{\hat{x}}^{t}_0 = f_{\theta}(\mathbf{p}_t, \mathbf{0}, t)$. Otherwise the model receives its previous prediction as an input: $\mathbf{\hat{x}}^{t}_0 = f_{\theta}(\mathbf{p}_t, \operatorname{SG}(\mathbf{\bar{x}}^t_0), t)$, where $\mathbf{\bar{x}}^{t}_0 = f_{\theta}(\mathbf{p}_t, \mathbf{0}, t)$ and $\operatorname{SG}$ is the stop-gradient function that prevent gradients from flowing through $\mathbf{\bar{x}}^t_0$. During the generation stage, the first prediction is made with self-condition set to zero and at all subsequent steps the predictions are performed as $\mathbf{\hat{x}}^{t}_0 = f_{\theta}(\mathbf{p}_t, \mathbf{\hat{x}}^{t+1}_0, t)$. We demonstrate the impact of self-conditioning in Appendix~\ref{sec:self-cond}.

\subsection{Sequence Length}
As diffusion models operate over fixed-length sequences, we pad all shorter sequences using a special padding token, which the model is trained to predict. In the end of generation padding tokens are discarded. To limit computational overhead, we set the maximum sequence length for each dataset to approximately the 99th percentile of training set sequence lengths. The exact values used for each dataset are provided in the Appendix \ref{sec:implementation}.

%% file: sources/experiments.tex
\section{Experiments}\label{sec:experiments}


\paragraph{Implementation Details}

In all experiments, we use a pre-trained embedding matrix, $\mathbf{E}$, from the BERT \citep{bert} model\footnote{We consider other embedding options in Appendix \ref{sec:embeddings}}. We normalize this matrix to have a zero mean and a unit variance and keep it fixed throughout training. Although the model receives the soft token distribution $\mathbf{p}_t$ as input, it does not operate directly on this distribution. Instead, we compute a weighted average of the token embeddings, $\mathbf{p}_t \mathbf{E}$, which yields a lower-dimensional, more tractable representation for the model to process.

Our model's architecture is based on the design proposed in \citet{tencdm}, consisting of Transformer decoder layers \citep{vaswani2017attention} augmented with UNet-style skip connections. Specifically, the output of the first layer is added to the input of the last, the second to the second-last, and so on. The full model has approximately 100M parameters. For conditional generation, we modify the model to accept an input sequence $\mathbf{w}^{x}$, which is processed by an additional 6-layer Transformer encoder. The encoder output is integrated into the decoder through cross-attention mechanisms. For timestep conditioning, we adopt the approach from \citet{diffuseq}, plugging learned timestep embeddings into each Transformer block akin to positional embeddings. The complete set of hyperparameters used for training and evaluation is provided in Appendix \ref{sec:implementation}.


\subsection{The Importance of $\tilde{\delta}$}\label{sec:delta}

\begin{figure}
  \begin{center}
    \includegraphics[width=0.8\linewidth]{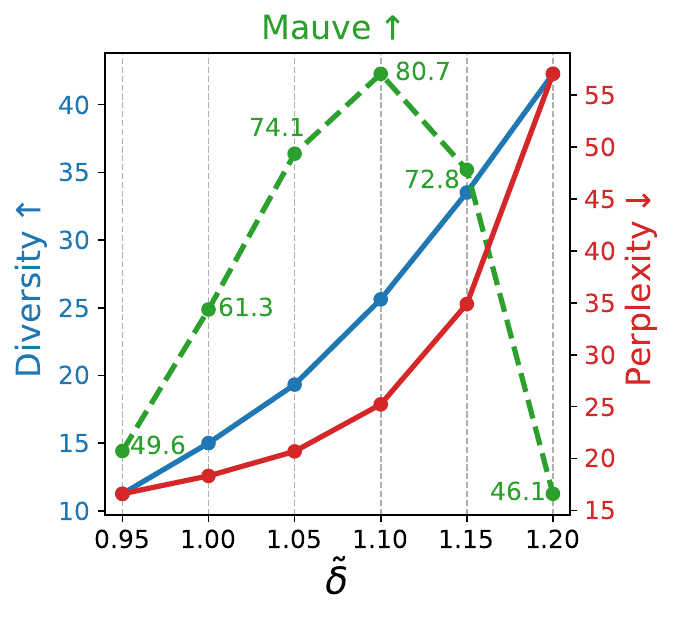}
  \end{center}
  \caption{Unconditional generation for $\delta = 1$ and varying $\tilde{\delta}$.}
  \label{fig:varying_delta}
\end{figure}

Before presenting results on seq-to-seq generation tasks, we highlight the importance of the hyperparameter $\tilde{\delta}$, which controls the stochasticity of the denoising process. To illustrate its impact, we evaluate generation quality on an \textit{unconditional} generation task using different values of $\tilde{\delta}$. Specifically, we use the \textbf{ROCStories} dataset and assess performance using three metrics: generative \textbf{perplexity} (to estimate average text quality), \textbf{diversity} (to measure lexical variety) \citep{su2022a}, and the \textbf{MAUVE Score} \citep{mauve} (to evaluate the overall similarity of generated texts to the reference distribution). When calculating MAUVE, we generate 1,000 texts five times with different seeds and compare them with 1,000 randomly sampled reference texts. We then average the results.

Figure~\ref{fig:varying_delta} shows the results for a model trained with $\delta = 1$. We observe that lower values of $\tilde{\delta}$ lead to better perplexity scores but lower diversity. In other words, reduced stochasticity improves the quality of individual texts but decreases their uniqueness. This trade-off is actually desirable for sequence-to-sequence tasks, where diversity typically arises naturally from the varying input conditions. In Appendix \ref{sec:seq2seq_delta}, we justify this insight by grid-searching the best $\tilde{\delta}$ value. As a result, we set $\tilde{\delta} = 0.1$ for all sequence-to-sequence experiments.

In contrast, for unconditional generation, the optimal value of $\tilde{\delta}$ is slightly higher than the one used during training, as indicated by the MAUVE Score. At this point, the generated texts exhibit sufficient diversity while maintaining acceptable perplexity. These findings show that $\tilde{\delta}$ has a strong influence on the generation process and should be tuned carefully depending on the target task.

\paragraph{Datasets}
In addition to the unconditional generation on \textbf{ROCStories} dataset, we evaluate \textsc{Smoothie} on four sequence-to-sequence datasets of varying difficulty. For \textit{paraphrase generation}, we use the Quora Question Pairs \textbf{(QQP)} dataset \citep{qqp}, which contains 147K pairs of semantically equivalent questions. For \textit{question generation}, we adopt the \textbf{Quasar-T} dataset \citep{quasar-t}, processed by \citet{diffuseq}, resulting in 119K document-question pairs. For \textit{summarization}, we use the \textbf{XSum} dataset \citep{xsum}, comprising 204K BBC articles and their corresponding summaries. For \textit{detoxification}, we use \textbf{ParaDetox} \citep{paradetox} dataset with 19,766 pairs of toxic and neutral comments. More detailed dataset information is provided in the Appendix \ref{sec:datasets}.

\paragraph{Metrics}
Following the evaluation protocol from \citet{diffuseq, tess}, we employ a combination of n-gram-based, diversity and semantic similarity metrics. Specifically, we report \textbf{BLEU} \citep{bleu} and \textbf{ROUGE-1/2/L} \citep{rouge} scores to measure lexical overlap between generated and reference texts, and \textbf{BERTScore (BS)} \citep{bertscore} to assess semantic similarity. For BERTScore, we use the \texttt{microsoft/deberta-xlarge-mnli} model to ensure consistency with previous studies \citep{seqdiffuseq, tess}.

To evaluate the diversity of generated texts, we compute n-gram diversity \citep{div}, which reports the fraction of unique unigrams (\textbf{Div-1}) and 4-grams (\textbf{Div-4}) in a text. Additionaly, for the text detoxification task, we measure \textbf{J-Score} \citep{j-score}, which comprises text fluency, style accuracy, and content preservation.

\paragraph{Baselines}
We compare \textsc{Smoothie} against several diffusion-based and autoregressive baselines, all with approximately 100M parameters and trained from scratch on each dataset. The diffusion-based baselines include DiffuSeq \citep{diffuseq}, SeqDiffuSeq \citep{seqdiffuseq}, AR-Diffusion \citep{ar-diffusion}, and GENIE \citep{genie}, SSD-LM \citep{han2022ssd}, TESS \citep{tess}, and TEncDM \citep{tencdm}. We also compare against MDLM \citep{mdlm}, an established masked diffusion model that we trained for sequence-to-sequence tasks using the provided code. For autoregressive baselines, we evaluate BART \citep{bart}, GPT-2 \citep{gpt-2}, GPVAE-T5 \citep{gpvae}, FLAN-T5 \citep{flan}, and a standard Transformer \citep{vaswani2017attention}. TESS approach uses pre-trained RoBERTa \citep{roberta} to initialize its diffusion model. For a fair comparison, we only compare to the model trained from random initialization.

\begin{table}
\fontsize{8}{9}\selectfont
\centering
\setlength{\tabcolsep}{4pt}
\caption{Generation time and peak memory consumption for \textsc{Smoothie}, embedding- and simplex-based diffusion.}
\label{tab:speed}
\begin{tabular}{l|cccc}
\toprule
\textbf{Method} & \textbf{Generation time (s)} & \textbf{Memory (MB)} \\
\midrule
Embedding & 1.642 & 593.3 \\
Simplex & 2.897 & 678.4 \\
\textsc{Smoothie} & 2.897 & 593.3 \\
\bottomrule
\end{tabular}
\end{table}

\begin{table*}[t]
\centering
\caption{Results on XSum (left) and Quasar-T (right) datasets. $\dagger$ denotes autoregressive models, $\triangle$ denotes the results reproduced with original code, $\star$ denotes our implementations. The best-performing \emph{diffusion} results are highlighted in \textbf{bold}, the second-best are \underline{underlined}.}
\label{tab:xsum_quasar}
\begin{minipage}[t]{.45\textwidth}
    \fontsize{8}{9}\selectfont
    \centering
    \renewcommand{\arraystretch}{1}
    \setlength{\tabcolsep}{6pt}
    \begin{tabular}{lcc}
    \toprule
    & \multicolumn{2}{c}{\textbf{XSum}} \\
    \cmidrule(r){2-3}
    \textbf{Method} & \textbf{BS} $\uparrow$ & \textbf{R-1/2/L} $\uparrow$ \\
    \midrule
    GPT-2$^{\dagger \triangle}$ & 69.0 & 28.3/8.2/21.8 \\
    Transformer$^\dagger$ & — & 30.5/10.4/24.2 \\
    FLAN-T5$^\dagger$ & 72.7 & 34.6/12.9/27.2 \\
    \midrule
    MDLM$^{\triangle}$ & 62.1 & 27.9/7.7/21.1 \\
    [0.25ex]\hdashline[1pt/1pt]\noalign{\vskip 0.5ex}
    DiffuSeq & 46.8 & 18.9/1.3/13.6 \\
    SeqDiffuSeq$^{\triangle}$ & 61.8 & 28.6/6.7/21.3 \\
    AR-Diffusion & — & 31.7/10.1/24.7 \\
    GENIE & — & 29.3/8.3/21.9 \\
    TEncDM & \textbf{69.9} & 31.9/\underline{10.7}/\underline{25.3} \\ 
    [0.25ex]\hdashline[1pt/1pt]\noalign{\vskip 0.5ex}
    Embedding$^\star$ & 68.2 & \underline{32.1}/10.1/24.6 \\
    Simplex$^\star$ & 63.8 & 29.6/8.5/23.0 \\
    \textsc{Smoothie}$^\star$ (ours) & \underline{68.8} & \textbf{33.7}/\textbf{11.1}/\textbf{26.0} \\
    \bottomrule
    \end{tabular}
\end{minipage}
\begin{minipage}[t]{.52\textwidth}
    \fontsize{8}{9}\selectfont
    \centering
    \renewcommand{\arraystretch}{0.95}
    \setlength{\tabcolsep}{3pt}
    \begin{tabular}{lcccc}
    \toprule
    & \multicolumn{4}{c}{\textbf{Quasar-T}} \\
    \cmidrule(r){2-5}
    \textbf{Method} & \textbf{BS} $\uparrow$ & \textbf{BLEU} $\uparrow$ & \textbf{R-L} $\uparrow$ & \textbf{D-1/4} \\
    \midrule
    GPT-2$^\dagger$ & 60.5 & 7.4 & 27.2 & 96.0/92.2 \\
    GPVAE-T5$^\dagger$ & 63.1 & 12.5 & 33.9 & 93.8/72.8 \\
    BART$^\dagger$ & 66.2 & 17.4 & 38.8 & 98.2/61.7 \\
    \midrule
    MDLM$^{\triangle}$ & 60.7 & 17.5 & 33.6 & 91.0/\textbf{64.2} \\
    [0.25ex]\hdashline[1pt/1pt]\noalign{\vskip 0.5ex}
    DiffuSeq & 59.4 & 15.8 & — & 91.1/— \\
    SeqDiffuSeq & 61.4 & 17.2 & — & 92.7/— \\
    SSD-LM & 62.8 & 14.1 & \textbf{38.5} & \underline{94.5}/56.9 \\
    TESS (random) & 60.8 & 19.0 & 36.1 & \textbf{96.1}/62.4  \\
    [0.25ex]\hdashline[1pt/1pt]\noalign{\vskip 0.5ex}    Embedding$^\star$ & 62.0 & 18.9 & 35.2 & 92.4/61.2 \\
    Simplex$^\star$ & \underline{63.0} & \underline{19.3} & \underline{36.9} & 93.0/\underline{63.8} \\
    \textsc{Smoothie}$^\star$ (ours) & \textbf{63.1} & \textbf{19.9} & 36.5 & 92.8/63.3 \\
    \bottomrule
    \end{tabular}
\end{minipage}
\end{table*}

Additionally, we conduct a rigorous comparison of our proposed distance-based latent space with two previously explored alternatives: the embedding space \citep{seqdiffuseq, diffuseq} (Embedding$^\star$ in experiments) and the simplex space \citep{han2022ssd, tess} (Simplex$^\star$ in experiments). To ensure a fair evaluation, we train all diffusion models under identical conditions, keeping the architecture, training hyperparameters, and decoding strategy fixed. The only variables are the latent space and its associated noise schedule. For embedding-based diffusion, we use the noise scheduler from \citet{tencdm}, while for simplex-based diffusion, we adopt the scheduler from \citet{han2022ssd}. In all three cases, sampling is performed using a procedure defined in the respective latent space, following the formulation in Eq.~\ref{eq:sampling}. \textsc{Smoothie} and the embedding-based diffusion model are trained using MSE loss, while the simplex-based diffusion is trained using cross-entropy loss because it is not suitable for predicting continuous embeddings.

\paragraph{Generation Speed}

During generation, \textsc{Smoothie} requires the calculation of pairwise distances between the predicted embeddings and all the embeddings in the vocabulary. This operation has a complexity of $\mathcal{O}(\text{batch size} \times \text{seq len} \times d \times V)$, which is the same as that of the linear head used in simplex diffusion \citep{tess, flm} or discrete diffusion models \citep{d3pm, mdlm, duo2} to predict tokens at each step. However, embedding-based diffusion does not have such operation and thus generates text faster\footnote{Some papers use a clamping trick \citep{diffusion-lm} that results in the same overall computational complexity as \textsc{Smoothie}.}.

We measure the difference in generation time and memory usage between the three approaches and present the results in Table \ref{tab:speed}. We performed 100 generation steps with a batch size of 32 and a sequence length of 80, reporting the total generation time and peak memory consumption. We observe that simplex diffusion and \textsc{Smoothie} have the same speed, while embedding diffusion is $1.75\times$ faster. To compensate for this, we increase the number of generation steps by $1.75\times$ for embedding diffusion compared to \textsc{Smoothie} in all our experiments.

\subsection{Empirical Results}

We now present a quantitative comparison of \textsc{Smoothie} against a range of generative models. Wherever possible, we adopt reported results from prior works \citep{ld4lg, ar-diffusion, tess, cosmos}. When certain metric values are unavailable, we reproduce the corresponding methods using the original implementations. For consistency, we re-implement and train the embedding- and simplex-based diffusion baselines within our framework. 

We show the results on XSum and Quasar-T dataset in Table~\ref{tab:xsum_quasar}, and on ROCStories, QQP, and ParaDetox in Table~\ref{tab:rocstories_qqp_paradetos}. In Appendix \ref{sec:openwebtext}, we also evaluate \textsc{Smoothie} of OpenWebText \citep{openwebtext}. Overall, \textsc{Smoothie} consistently outperforms other text diffusion approaches, as well as diffusion methods based on embedding- and simplex-based latent spaces achieving quality comparable to that of autoregressive models.

Notably, embedding-based diffusion performs better than simplex-based diffusion on all datasets except Quasar-T. This difference can be attributed to the fact that simplex-based diffusion does not incorporate semantic information into the noising process, making it inherently more chaotic. Nevertheless, when combined with our proposed architecture, simplex-based diffusion surpasses the TESS approach, which employs the same diffusion process and a training pipeline, differing only in the architecture design. This highlights that selecting an appropriate model architecture is as critical as choosing the diffusion space.

The most pronounced improvement in generation quality is observed on the ROCStories dataset. By tuning the $\tilde{\delta}$ parameter (Section~\ref{sec:delta}), \textsc{Smoothie} effectively balances diversity and coherence, achieving the highest MAUVE score and nearly matching the quality of GPT-2.

\begin{table*}
\fontsize{8}{9}\selectfont
\caption{Text generation results on ROCStories, QQP and ParaDetox datasets. $\dagger$ denotes autoregressive models, $\triangle$ denotes the results reproduced with original code, $\star$ denotes our implementations. The best-performing \emph{diffusion} results are highlighted in \textbf{bold}, the second-best are \underline{underlined}.}
\label{tab:rocstories_qqp_paradetos}
\centering
\renewcommand{\arraystretch}{1}
\setlength{\tabcolsep}{4pt}
\begin{tabular}{lccc|cccc|cc}
\toprule
& \multicolumn{3}{c}{\textbf{ROCStories}} & \multicolumn{4}{c}{\textbf{QQP}} & \multicolumn{2}{c}{\textbf{ParaDetox}} \\
\textbf{Method} & \textbf{MAUVE} $\uparrow$ & \textbf{PPL} $\downarrow$ & \textbf{Div} $\uparrow$ & \textbf{BS} $\uparrow$ & \textbf{BLEU} $\uparrow$ & \textbf{R-L} $\uparrow$ & \textbf{D-1/4} $\uparrow$ & \textbf{BLEU} $\uparrow$ & \textbf{J-Score} $\uparrow$ \\
\midrule
GPT-2$^\dagger$ & 78.9 & 20.5 & 25.2 & 82.5 & 19.8 & 52.1 & 98.0/62.5 & 67.7 & 60.4 \\
GPVAE-T5$^\dagger$ & — & — & — & 84.7 & 24.1 & 58.9 & 96.9/61.7 & — & — \\
BART$^\dagger$ & — & — & — & 85.7 & 30.4 & 61.4 & 98.8/61.0 & — & — \\
\midrule
MDLM$^{\triangle}$ & 63.9 & 58.1 & \textbf{35.1} & 76.3 & 21.5 & 46.2 & 96.2/64.4 & 61.5 & 41.4 \\
[0.25ex]\hdashline[1pt/1pt]\noalign{\vskip 0.5ex}
DiffuSeq & 8.6 & 50.5 & 12.4 & 79.5 & 18.5 & — & 97.6/— & 67.9 & 47.5 \\
SeqDiffuSeq & 10.3 & 29.3 & 13.7 & 82.9 & 23.3 & — & 98.1/— & \underline{68.8} & 48.6 \\
AR-Diffsion$^{\triangle}$ & 6.6 & 41.8 & 10.1 & 80.1 & 19.2 & 54.9 & — & 64.7 & 46.5 \\
SSD-LM & — & — & — & \underline{83.8} & 22.9 & \underline{58.3} & \textbf{98.8}/57.3 & — & — \\
TEncDM & \underline{76.2} & 29.1 & \underline{29.5} & \underline{83.8} & 30.7 & 57.3 & — & 61.9 & \underline{49.6} \\
[0.25ex]\hdashline[1pt/1pt]\noalign{\vskip 0.5ex}
Embedding$^\star$ & 41.5 & 28.3 & 26.1 & 83.4 & \textbf{31.3} & 59.4 & 97.7/\underline{64.5} & 67.6 & 49.1 \\
Simplex$^\star$ & 15.2 & \underline{25.3} & 12.4 & 80.6 & 26.8 & 54.9 & 96.8/\textbf{64.8} & 65.1 & 47.7 \\
\textsc{Smoothie}$^\star$ (ours) & \textbf{80.7} & \textbf{24.9} & 25.1 & \textbf{83.9} & \underline{30.8} & \textbf{60.9} & \underline{98.4}/60.5 & \textbf{69.2} & \textbf{51.7} \\
\bottomrule
\end{tabular}
\end{table*}

\subsection{Amount of Denoising Steps}\label{sec:steps}

\begin{table}
\fontsize{8}{9}\selectfont
\centering
\setlength{\tabcolsep}{4pt}
\caption{The impact of changing the number of steps on generation quality. We show J-score for ParaDetox and BERTScore for the other datasets.}
\label{tab:lengths}
\begin{tabular}{l|cccc}
\toprule
\textbf{Steps}& \textbf{XSum} & \textbf{Quasar-T} & \textbf{QQP} & \textbf{ParaDetox} \\
\midrule
25 & 67.7 & \textbf{63.1} & \textbf{83.9} & 51.1 \\
50 & 68.5 & \textbf{63.1} & 83.8 & 51.4 \\
100 & 68.7 & \textbf{63.1} & 83.7 & \textbf{51.7} \\
200 & \textbf{68.8} & \textbf{63.1} & 83.6 & 51.0 \\
500 & 68.4 & \textbf{63.1} & 83.5 & 50.8 \\
\bottomrule
\end{tabular}
\end{table}

Table~\ref{tab:lengths} presents the relationship between the number of denoising steps and the generation quality of \textsc{Smoothie} in terms of J-Score for ParaDetox and BERTScore for other datasets. We observe that for all datasets except ParaDetox, the quality does not change much regardless of the number of steps. Nevertheless, for XSum the performance improves as the number of steps increases until we reach 200 steps, after which the quality drops. This can be explained by the impact of self-conditioning, which lead to a mismatch between train and generation trajectory for larger amount of steps \cite{tencdm}. Overall, the results align with the observation made in the TESS paper~\citep{tess}, which suggests that the optimal number of denoising steps correlates with the complexity of the task.

\subsection{Diffusion Solver}\label{sec:solver}

In all experiments, we use the solver defined in Eq. \ref{eq:sampling} rather than the commonly used DDPM \citep{ddpm} or DDIM \citep{ddim} solvers. This choice is motivated by the consistently superior empirical performance of our solver. To enable a comparison with DDPM and DDIM, we adapt the originally proposed forward process (Eq. \ref{eq:forward}) by introducing a timestep-dependent variance.
\begin{align}
q(\mathbf{D}_t | \mathbf{D}_0) = \mathcal{N}\left(\mathbf{D}_t \bigg| \frac{1}{\sigma_t^2}\mathbf{D}_0, \left(1 - \frac{1}{\sigma_t^4}\right)\delta^2 I\right)
\end{align}
A quantitative comparison between DDPM, DDIM and our solver is reported in Table \ref{tab:solver}. The results demonstrate a marginal yet consistent advantage of our solver across all evaluated tasks. We hypothesize that this behavior stems from the discrete nature of text embeddings, which leads to a highly sparse data distribution. Such sparsity may be unfavorable for both the DDPM and DDIM solvers, as they move the lantent in a predicted direction and even a small deviation in this direction can cause the latent representation to leave the data manifold, which will cause more inaccuracy in further predictions.

In contrast, our solver does not rely on the previous latent state when computing updates. Instead, it directly applies the forward process to the model's prediction. Although this strategy may appear suboptimal, it reduces the risk of drifting in incorrect directions, which is particularly important for sparse distributions. Nevertheless, a deeper theoretical understanding of the observed performance gap remains an open question and is left for future work.

\begin{table}
\fontsize{8}{9}\selectfont
\centering
\setlength{\tabcolsep}{4pt}
\caption{Generation quality of DDPM, DDIM and our (Eq. \ref{eq:sampling}) diffusion solvers on Seq2seq tasks.}
\label{tab:solver}
\begin{tabular}{l|cc|cc|cc|c}
\toprule
& \multicolumn{2}{c}{\textbf{XSum}} & \multicolumn{2}{c}{\textbf{Quasar-T}} & \multicolumn{2}{c}{\textbf{QQP}} & \textbf{ParaDetox} \\

\textbf{Solver} & \textbf{BS} $\uparrow$ & \textbf{R-L} $\uparrow$ & \textbf{BS} $\uparrow$ & \textbf{R-L} $\uparrow$ & \textbf{BS} $\uparrow$ & \textbf{R-L} $\uparrow$ & \textbf{J-Score} $\uparrow$ \\
\midrule
DDPM & 68.5 & 25.6 & 62.8 & 35.5 & 83.7 & 60.7 & 51.5 \\
DDIM & 67.4 & 24.6 & 60.3 & 31.8 & 82.0 & 57.9 & 50.2 \\
Ours & \textbf{68.8} & \textbf{26.0} & \textbf{63.1} & \textbf{36.5} & \textbf{83.9} & \textbf{60.9} & \textbf{51.7} \\
\bottomrule
\end{tabular}
\end{table}

%% file: sources/limitations.tex
\section{Limitations}\label{sec:limitations}

\paragraph{Pre-trained Embeddings}
\textsc{Smoothie} relies on a pre-trained embedding matrix $\mathbf{E}$ from the BERT model. While this choice simplifies the training process and improves its stability, it may hold back the model's generation quality. Finetuning embeddings for a specific task should offer better results. Also, an end-to-end training approach, as used in \citet{diffusion-lm, diffuseq}, could be applied to our method. However, training embeddings simultaneously with the diffusion model is hard to stabilize, because embeddings collapse when diffusion model is optimized only with MSE between text embeddings and model's predictions. Addition of a cross-entropy loss prevents the collapse, but might lead to the explosion of embeddings' norm, again making the task trivial for the diffusion model. For these reasons, we leave the exploration of an end-to-end training approach for future work.

\paragraph{Complexity of the Distance Matrix Computation}

\textsc{Smoothie} requires computing a matrix of pairwise distances between the predicted embedding and all vocabulary embeddings at every generation step, which slows down sampling. While several related approaches such as SSD-LM \citep{han2022ssd}, TESS \citep{tess}, and Diffusion-LM \citep{diffusion-lm} have the same sampling complexity, alleviating this bottleneck is an important direction for improvement.

Unfortunately, this issue cannot be easily addressed by straightforward approximations, such as restricting the computation to top-$k$ nearest embeddings. At most timesteps, nearly all vocabulary embeddings receive non-zero weights $\mathbf{p}_t$, making such approximation inaccurate. In principle, this limitation could be mitigated by clustering embeddings at each timestep and performing smoothing using cluster centroids, or by reducing the effective vocabulary size. However, both of these methods are very complex and go beyond the scope of this work.


%% file: sources/conclusion.tex
\section{Conclusion}

In this work, we introduce \textsc{Smoothie}, a text diffusion method that constructs its diffusion process with consideration of the discrete nature of text and the semantic relationships between tokens. To capture these properties, each token is mapped to a vector of Euclidean distances between its embedding and the embeddings of all tokens in the vocabulary. Our choice of the Euclidean distance is based on the Euclidean semantic space hypothesis \citep{euclidean_space_hyp}, which posits that semantic similarity correlates with Euclidean proximity in embedding space. Additionally, we use a sampling method that disregards the previous latent state when calculating the updated latent. This solver demonstrates superior empirical performance to the conventionally used DDPM.

Our method also can be applicable to other categorical domains where semantic relationships exist between categories (e.g. graphs, protein sequences). However, in such cases, a different distance metric more suited to the domain's properties may be required. We leave the exploration of this direction to future work.

Empirical results on four sequence-to-sequence tasks demonstrate that \textsc{Smoothie} outperforms existing text diffusion methods, as well as our diffusion model framework with alternative diffusion latent spaces that do not rely on additional encoders.

%% file: sources/acknowledgments.tex
\section*{Acknowledgments}

We are grateful to Ildus Sadrtdinov for his valuable insights and discussions throughout this project. The paper was supported in part through computational resources of HPC facilities at HSE University.

%% file: sources/societal_impact.tex
\section*{Impact Statement}\label{sec:societal_impact}

This paper presents work whose goal is to advance the field of Natural Language Processing. There are many potential societal consequences of our work, none which we feel must be specifically highlighted here.

%% file: sources/appendix.tex


\section{Relationship Between Distance-based and Simplex-based Latent Spaces}\label{sec:smoothie_tess}

In this section, we demonstrate that our proposed \emph{distance-based latent space} generalizes the \emph{simplex-based latent space}. Specifically, we show that the simplex-based latent space corresponds to a special case of a distance-based latent space when equipped with a trivial metric.

\textsc{Smoothie} maps each token \( w \) to a latent vector \( \mathbf{d}^w \), where each component is given by:
\begin{align}
\mathbf{d}_{(i)}^w = -\frac{1}{2} \|\mathbf{E}_w - \mathbf{E}_i\|^2.
\end{align}

For other categorical domains, the Euclidean distance can be replaced with a more suitable metric \( \rho(w, i) \), leading to:
\begin{align}
\mathbf{d}_{(i)}^w = -\rho(w, i).
\end{align}

To relate this to simplex-based representations, consider the case where \( \rho \) is the \emph{trivial metric}:
\begin{align}
\rho(w, i) = [w \neq i],
\end{align}

i.e., \( 0 \) when \( w = i \) and \( 1 \) otherwise. Under this choice, the latent vector becomes:
\begin{align}
\mathbf{d}_{(i)}^w = 
\begin{cases}
    0, & i = w, \\
    -1, & \text{otherwise}.
\end{cases}
\end{align}

In comparison, the simplex-based latent space maps each token \( w \) to a vector \( \mathbf{s}^w \) in the \( k \)-logit simplex:
\begin{align}
\mathbf{s}_{(i)}^w = 
\begin{cases}
    +k, & i = w, \\
    -k, & \text{otherwise}.
\end{cases}
\end{align}

Both \textsc{Smoothie} and simplex diffusion apply a Gaussian diffusion process to corrupt the latent vector:
\begin{align}
\mathbf{z}_t = \phi_t \mathbf{z}_0 + \gamma_t \varepsilon,
\end{align}

where \(\mathbf{z}_0 \in \{\mathbf{d}^w, \mathbf{s}^w\} \) and \( \varepsilon \sim \mathcal{N}(0, I) \). To form a model input, the corrupted vector is then transformed into a probability distribution using the softmax function:
\begin{align}
p_t = \operatorname{softmax}(\mathbf{z}_t).
\end{align}

Since the softmax function is invariant to uniform additive shifts, we have:
\begin{align}
\operatorname{softmax}(\phi_t \mathbf{s}^w + \gamma_t \varepsilon) 
= \operatorname{softmax}(\phi_t (\mathbf{s}^w - k) + \gamma_t \varepsilon) = \operatorname{softmax}(2k\phi_t \mathbf{d}^w + \gamma_t \varepsilon),
\end{align}
where the final equality follows from observing that \( \mathbf{s}^w - k = 2k \mathbf{d}^w \).

This confirms that the simplex-based latent space is equivalent, up to scaling, to the distance-based latent space under the trivial metric. Hence, the simplex-based representation is a special case within the more general distance-based latent space framework.

\section{Derivation of the Loss Function}\label{sec:loss_function}

\begin{align}
    & -\log p_\theta(\mathbf{D}_{0}) = -\log \int \frac{p_\theta(\mathbf{D}_{0:T}) q(\mathbf{D}_{1:T} | \mathbf{D}_0)}{q(\mathbf{D}_{1:T} | \mathbf{D}_0)} \mathrm{d} \mathbf{D}_{1:T} \le -\mathbb{E}_q \log \frac{p_\theta(\mathbf{D}_{0:T})}{q(\mathbf{D}_{1:T} | \mathbf{D}_{0})} \\
    & = -\mathbb{E}_q \left[\log \frac{p_\theta(\mathbf{D}_T)}{q(\mathbf{D}_T | \mathbf{D}_0)} + \sum_{t=2}^T \log \frac{p_\theta(\mathbf{D}_{t-1} | \mathbf{D}_{t})}{q(\mathbf{D}_{t-1} | \mathbf{D}_{0})} + \log p_\theta(\mathbf{D}_0 | \mathbf{D}_1) \right] \\
    & = \mathbb{E}_q \bigg[\underbrace{\mathrm{D_{KL}}\big[q(\mathbf{D}_T | \mathbf{D}_0) \| p(\mathbf{D}_T)\big]}_{L_T} + \sum_{t=2}^T \underbrace{\mathrm{D_{KL}}\big[q(\mathbf{D}_{t-1} | \mathbf{D}_0) \| p_\theta(\mathbf{D}_{t-1} | \mathbf{D}_t)\big]}_{L_{t-1}} \underbrace{- \log p_\theta(\mathbf{D}_0 | \mathbf{D}_1)}_{L_0}\bigg]
\end{align}

In this formula, $L_T$ is constant during the training, as it does not depend on any learnable parameters. Both forward and reverse processes are defined by Gaussian distributions, which allows us to compute the terms $L_0$ and $L_{t-1}$ in closed form:
\begin{align}
&L_0 = \mathbb{E}_q \left[ \frac{1}{2\tilde{\delta}^2} \left\|\mathbf{D}_0 - \mu_{\theta}(\mathbf{p}_1, 1)\right\|^2 \right] + C_0; \;\; L_{t-1} = \mathbb{E}_q \left[\frac{1}{2\tilde{\delta}^2} \left\|\frac{1}{\sigma_t^2} \mathbf{D}_0 - \mu_{\theta}(\mathbf{p}_t, t)\right\|^2\right] + C_{t-1},
\end{align}

where $C_0$ and $C_{t-1}$ are constants that do not depend on parameters $\theta$. This implies that the most direct parameterization of $\mu_{\theta}$ is a model that predicts $\mathbf{D}_0 / \sigma_t^2$, corresponding to the posterior mean of the forward process. However, for practical reasons, we instead parameterize $\mu_{\theta}$ as $g_{\theta} / \sigma_t^2$ which ensures that all model outputs are scaled to have the same variance across timesteps.
\begin{align}
L_{t-1} = \mathbb{E}_q \left[\frac{1}{2\tilde{\delta}^2\sigma_t^4} \left\|\mathbf{D}_0 - g_{\theta}(\mathbf{p}_t, t)\right\|^2\right] + C_{t-1},
\end{align}

Following \citet{ddpm}, we replace $L_{t-1}$ with its simplified version by removing the scaling coefficient $2\tilde{\delta}^2\sigma_t^4$, resulting in the following loss function:
\begin{align}
L_{\mathbf{D}}(\theta) = \mathbb{E}_{\mathbf{w}^y, t, \mathbf{p}_t} \left[\|\mathbf{D}_0(\mathbf{E}_{\mathbf{w}^y}) - g_{\theta}(\mathbf{p}_t, t)\|^2 \right]
\end{align}

\section{Proof of Theorem 4.1}\label{sec:proof_4.1}

\begin{proof}
We begin by recalling a standard result:

\textbf{Lemma.}
The minimum value of the function \(\mathbb{E}_{\mathbf{y}}\left[\|\mathbf{y} - \mathbf{z}\|^2\right]\) is achieved when \(\mathbf{z} = \mathbb{E}[\mathbf{y}]\).

Using this lemma, we obtain:
\begin{align}
g^*(\mathbf{p}_t, t) = \mathbb{E}_{\mathbf{w}^y}[\mathbf{D}_0(\mathbf{E}_{\mathbf{w}^y})] = \mathbb{E}_{\mathbf{w}^y}\Big[-\frac{1}{2}\big\{\|\mathbf{E}_{w_i^y} - \mathbf{E}_{j}\|^2\big\}_{i,j=1}^{m,V}\Big] \quad \text{and} \quad f^*(\mathbf{p}_t, t) = \mathbb{E}_{\mathbf{w}^y}[\mathbf{E}_{\mathbf{w}^y}],
\end{align}

where $\mathbf{w}^y \sim p(\mathbf{w}^y \mid \mathbf{p}_t)$. Since both $g^*(\mathbf{p}_t, t)$ and $f^*(\mathbf{p}_t, t)$ are matrices, without loss of generality we will prove this statement for an arbitrary row $i$ and column $j$. For brevity, we will define $u = \mathbf{E}_{w_i^y}$ and $v = \mathbf{E}_{j}$. Then, we need to show that

\begin{align}
\mathbb{E}_u\Big[-\frac{1}{2}\|u - v\|^2\Big] = -\frac{1}{2}\|\mathbb{E}[u] - v\|^2 + C
\end{align}

Expanding both sides:
\begin{align*}
\mathbb{E}_u \left[ \| u - v \|^2 \right] &= \mathbb{E}[\|u\|^2] - 2 v^\top \mathbb{E}[u] + \|v\|^2 \\
\| \mathbb{E}[u] - v \|^2 &= \|\mathbb{E}[u]\|^2 - 2 v^\top \mathbb{E}[u] + \|v\|^2
\end{align*}

Subtracting:
\[
\mathbb{E}[\|u\|^2] - \|\mathbb{E}[u]\|^2 = \sum_{k=1}^d \operatorname{Var}(u_k) = C
\]

Thus,
\[
\mathbb{E}_u \left[ -\frac{1}{2} \| u - v \|^2 \right] = -\frac{1}{2} \| \mathbb{E}[u] - v \|^2 - \underbrace{\frac{1}{2} C}_{\text{constant}},
\]
where \(C\) is a constant independent of \(\mathbb{E}[u]\).

Since this holds for all \( (i,j) \), the matrix identity holds:
\[
g^*(\mathbf{p}_t, t) = \mathbf{D}_0(f^*(\mathbf{p}_t, t)) + \mathbf{C}
\]
\end{proof}

\section{OpenWebText results}\label{sec:openwebtext}

\begin{wraptable}{r}{0.45\textwidth}
\fontsize{8}{9}\selectfont
\centering
\setlength{\tabcolsep}{4pt}
\caption{Text generation results on the OpenWebText dataset. $\triangle$ denotes the results reproduced with original code, $\star$ denotes our implementations.}
\label{tab:owt_results}
\begin{tabular}{l|ccc}
\toprule
\textbf{Method} & \textbf{MAUVE} $\uparrow$ & \textbf{PPL} $\downarrow$ & \textbf{Div} $\uparrow$ \\
\midrule
GIDD$^{\triangle}$ \citep{gidd} & 28.6 & 228.3 & \textbf{58.8} \\ 
Embedding$^{\star}$ & 55.7 & 51.9 & 21.2 \\
\textsc{Smoothie}$^{\star}$ ($\tilde{\delta} = 0.96$) & 58.9 & \textbf{48.2} & 19.8 \\
\textsc{Smoothie}$^{\star}$ ($\tilde{\delta} = 1.02$) & \textbf{59.9} & 62.2 & 23.4 \\
\bottomrule
\end{tabular}
\end{wraptable}

In this section, we provide the numerical results for \textsc{Smoothie}, embedding-based diffusion and GIDD, a strong masked diffusion model \citep{gidd} on the large scale OpenWebText dataset \citep{openwebtext} with the sequence length on 512 tokens. We use DDPM solver for embedding diffusion, and for \textsc{Smoothie} we perform 40\% of first steps with DDPM and switch to the solver in Eq. \ref{eq:sampling} for the remaining steps. We run \textsc{Smoothie} for $100$ steps, and, to align the methods in terms of computational complexity, we use $1.75\times$ more steps for embedding diffusion. Table \ref{tab:owt_results} shows that \textsc{Smoothie} significantly outperforms GIDD and marginally surpasses embedding diffusion by MAUVE metric.

\section{An Impact of $\tilde{\delta}$ on Seq2seq Tasks}\label{sec:seq2seq_delta}

\begin{wraptable}{r}{0.4\textwidth}
\vspace{-0.5cm}
\fontsize{8}{9}\selectfont
\centering
\setlength{\tabcolsep}{4pt}
\caption{The impact of $\tilde{\delta}$ value on generation quality. We show J-Score for ParaDetox and BERTScore for the other datasets.}
\label{tab:seq2seq_delta}
\begin{tabular}{l|cccc}
\toprule
\textbf{$\tilde{\delta}$} & \textbf{XSum} & \textbf{Quasar-T} & \textbf{QQP} & \textbf{ParaDetox} \\
\midrule
0.1 & \textbf{68.8} & \textbf{63.1} & \textbf{83.7} & \textbf{51.7} \\
0.25 & 68.7 & \textbf{63.1} & \textbf{83.7} & 51.4 \\
0.5 & 68.7 & \textbf{63.1} & \textbf{83.7} & 51.3 \\
0.75 & 68.6 & \textbf{63.1} & 83.6 & 50.9 \\
1 & 68.2 & \textbf{63.1} & 83.4 & 50.9 \\
\bottomrule
\end{tabular}
\end{wraptable}

In this section, we measure how the quality of sequence-to-sequence generation changes when the value of $\tilde{\delta}$ varies. For this experiment, we consider values in the range of $\tilde{\delta} \in \{0.1, 0.25, 0.5, 0.75, 1\}$ and set the number of generation steps to $100$. Table \ref{tab:seq2seq_delta} reports J-Score for ParaDetox and BERTScore for all other datasets. Although the difference in quality for different $\tilde{\delta}$ is not as significant as for the unconditional generation, it can be seen that lower values of $\tilde{\delta}$ produce better quality overall. Following these results, we set $\tilde{\delta} = 0.1$ for all datasets.

\section{Embeddings Ablation}\label{sec:embeddings}

\begin{wraptable}{r}{0.4\textwidth}
\vspace{-0.5cm}
\centering
\setlength{\tabcolsep}{4pt}
\caption{The generation quality of \textsc{Smoothie} trained with different embedding types on ROCStories dataset.}
\label{tab:embeddings}
\begin{tabular}{l|ccc}
\toprule
Embeddings & \textbf{MAUVE} $\uparrow$ & \textbf{PPL} $\downarrow$ & \textbf{Div} $\uparrow$ \\
\midrule
BERT (default) & 80.7 & 24.9 & 25.1 \\
GPT-2 & 64.4 & 23.1 & 25.0 \\
GloVe & 36.8 & 36.4 & 24.6 \\
\bottomrule
\end{tabular}
\end{wraptable}

Throughout this work, we utilize BERT embedding matrix to represent text tokens without additional comments. We find it important to evaluate the robustness of \textsc{Smoothie} to other choices of embeddings. Therefore, we demonstrate how model performance changes on the ROCStories dataset when embeddings are changed. We choose two alternatives with the same hidden size: GPT-2 \citep{gpt-2} embeddings with the vocabulary size of 50k and GloVe \citep{glove} embeddings trained manually on Wikipedia dataset for BPE tokens with the vocabulary size of 10k. In the Table \ref{tab:embeddings} we show the results of the ablation.

\begin{wraptable}{r}{0.3\textwidth}
\vspace{-0.5cm}
\centering
\setlength{\tabcolsep}{4pt}
\caption{An impact of the parameter $d$ in noise scheduler on the generation quality on the ROCStories dataset.}
\label{tab:noise_scheduler}
\begin{tabular}{l|ccc}
\toprule
& \textbf{MAUVE} $\uparrow$ & \textbf{PPL} $\downarrow$ & \textbf{Div} $\uparrow$ \\
\midrule
$d=4$ & 74.2 & 24.4 & 24.5 \\
$d=5$ & 80.7 & 24.9 & 25.1 \\
$d=6$ & 76.5 & 26.7 & 27.7 \\
$d=7$ & 68.9 & 24.6 & 26.7 \\
\bottomrule
\end{tabular}
\end{wraptable}

In terms of perplexity and diversity, GPT-2 embeddings perform similarly to BERT, with the exception of MAUVE. However, these results are still better than those of most other methods (see Table \ref{tab:rocstories_qqp_paradetos}). Interestingly, we found out that the optimal value of $\tilde{\delta}$ for GPT2 embeddings is lower than for BERT embeddings (1.03 vs 1.1). Most probably, this is because diversity increases naturally with the increase of the vocabulary size and the need to increase it artificially disappears. GloVe embeddings are worse than the ones extracted from a language model. Therefore, a significant drop in quality is not surprising. We can conclude that embeddings is an important component of the framework and the quality of the model does depend on the quality of embeddings. However, the method allows for flexibility in the choice of embeddings, which improves its applicability.

\section{Self-conditioning}\label{sec:self-cond}

Previous studies have shown that self-conditioning significantly improves the quality of text diffusion models \citep{seqdiffuseq, tencdm, tess, cdcd}. In this section, we compare the performance of \textsc{Smoothie}, as well as of embedding- and simplex-based diffusion models, with and without self-conditioning. The results on the XSum, Quasar-T, and QQP datasets are reported in Table \ref{tab:self-conditioning}. Although performance gains vary across models and datasets, self-conditioning generally improves quality, which confirms the previous observations.

\begin{table}[]
\caption{Impact of self-conditioning on the generation performance on XSum, Quasar-T and QQP datasets.}
\label{tab:self-conditioning}
\centering
\renewcommand{\arraystretch}{1}
\setlength{\tabcolsep}{6pt}
\begin{tabular}{lcc|ccc|ccc}
\toprule
& \multicolumn{2}{c}{\textbf{XSum}} & \multicolumn{3}{c}{\textbf{Quasar-T}} & \multicolumn{3}{c}{\textbf{QQP}} \\
\cmidrule(r){2-3} \cmidrule(r){4-6} \cmidrule(r){7-9}
\textbf{Method} & \textbf{BS} $\uparrow$ & \textbf{R-L} $\uparrow$ & \textbf{BS} $\uparrow$ & \textbf{BLEU} $\uparrow$ & \textbf{R-L} $\uparrow$ & \textbf{BS} $\uparrow$ & \textbf{BLEU} $\uparrow$ & \textbf{R-L} $\uparrow$ \\
\midrule
Embedding & 68.2 & 24.6 & 62.0 & 18.9 & 35.2 & 83.5 & 31.6 & 59.6 \\
\quad\; w/o SC & 65.2 & 23.6 & 62.9 & 19.5 & 36.0 & 81.7 & 27.7 & 57.4 \\
[0.25ex]\hdashline[1pt/1pt]\noalign{\vskip 0.5ex}
Simplex & 63.8 & 23.0 & 63.0 & 19.3 & 36.9 & 81.2 & 27.3 & 55.0 \\
\quad\; w/o SC & 61.2 & 21.5 & 62.5 & 19.4 & 36.4 & 80.0 & 25.9 & 54.1 \\
[0.25ex]\hdashline[1pt/1pt]\noalign{\vskip 0.5ex}
\textsc{Smoothie} & 68.8 & 26.0 & 63.0 & 19.0 & 35.8 & 83.9 & 30.8 & 60.9 \\
\quad\; w/o SC & 67.5 & 25.4 & 61.9 & 19.0 & 35.7 & 83.2 & 29.4 & 59.9 \\
\bottomrule
\end{tabular}
\end{table}

\section{Noise Scheduler Ablation}\label{sec:noise_scheduler}

\begin{wrapfigure}{r}{0.5\textwidth}
  \vspace{-0.8cm}
  \begin{center}
    \includegraphics[width=1\linewidth]{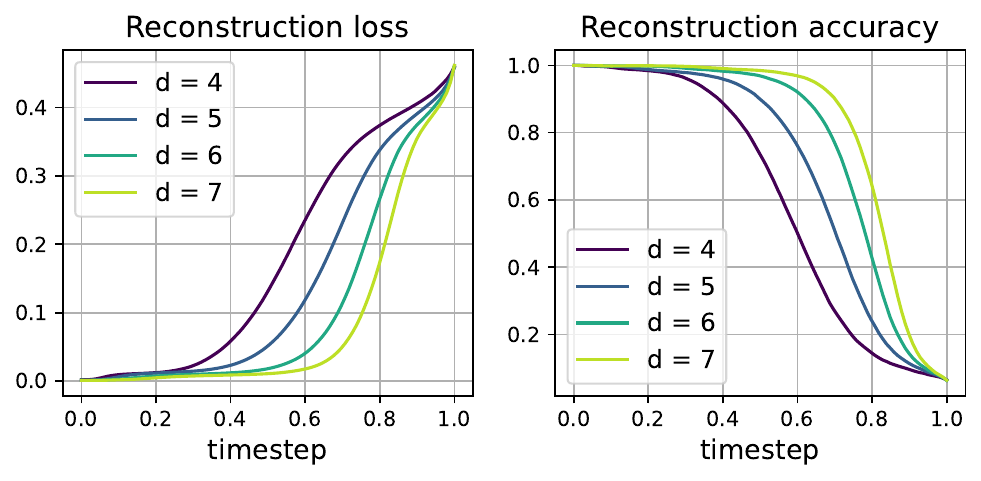}
  \end{center}
  \caption{Reconstruction loss (left) and reconstruction accuracy (right) w.r.t. timestep for \textsc{Smoothie}, trained with \emph{arctan} noise scheduler with $d=5$.}
  \label{fig:per_t_stats}
\end{wrapfigure}

In this work, we use a special \emph{arctan} noise scheduler for \textsc{Smoothie} to make sure that the model entropy grows linearly with $t$ \citep{cdcd}. In this section, we perform an ablation study for the proposed noise scheduler by evaluating different values of $d$. In Table \ref{tab:noise_scheduler}, we show the numerical performance on the ROCStories dataset. For each 
$d$ we chose the best $\tilde{\delta}$ based on MAUVE. Smaller $d$ values correspond to more aggressive corruption. The results suggest that while the difference is marginal, \textsc{Smoothie} is sensitive to the choice of the noise scheduler.

Figure \ref{fig:per_t_stats} illustrates how the reconstruction loss and the accuracy of the predicted tokens depend on the timestep t for our noise scheduler. In other words, we evaluate how closely the prediction $\mathbf{\hat{x}}_0 = f_{\theta}(\mathbf{p}_t, \mathbf{0}, t)$ matches the original $\mathbf{x}_0$. Accuracy is calculated only for non-padding tokens.

\section{Training Dynamics}\label{sec:training_dynamics}

\begin{wrapfigure}{r}{0.27\textwidth}
\vspace{-0.9cm}
\begin{center}
\includegraphics[width=1\linewidth]{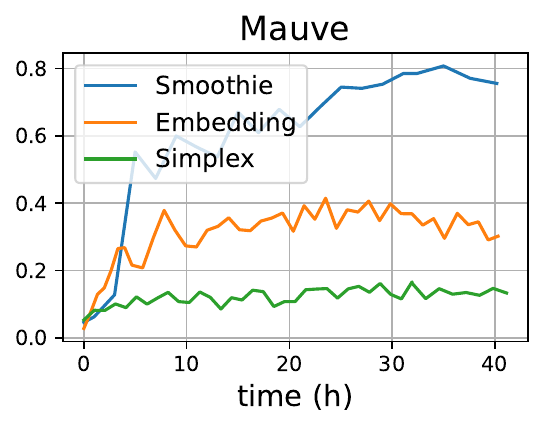}
\end{center}
\caption{Training dynamics of \textsc{Smoothie}, embedding and simplex diffusions on the ROCStories dataset.}
\label{fig:training_dynamics}
\end{wrapfigure}

In this section, we examine the differences in the training dynamics of \textsc{Smoothie}, embedding and simplex diffusions on the ROCStories dataset. Figure \ref{fig:training_dynamics} illustrates how the Mauve score changes with respect to training time. For embedding diffusion, we perform $1.75\times$ more generation steps (350 vs 200) than for the other diffusion types to match the generation time. The results suggest that, although \textsc{Smoothie} trains and generates text more slowly than the embedding diffusion, it significantly outperforms other methods throughout the training process.

\section{Implementation Details}\label{sec:implementation}

The hyperparemeters for training and inference of the models across all datasets are presented in Table \ref{tab:hyperameters}. We trained our models using two 80 GB NVIDIA A100 GPUs for 15 hours on average. For all the tasks, we save checkpoints every 25,000 steps. We select the best checkpoint by the quality on the development set. During generation we do not apply the clamping trick \citep{diffusion-lm}, since it does not improve quality in our experiments. We do not use the classifier-free guidance \citep{ho2021classifierfree} for the same reason.

\begin{table}
\centering
\caption{Complete hyperparameter configurations for all datasets.}
\label{tab:hyperameters}
\begin{tabular}{lccccc}
\toprule
Hyperparameter & \textbf{ROCStories} & \textbf{XSum} & \textbf{Quasar-T} & \textbf{QQP} & \textbf{ParaDetox} \\
\midrule
Tokenizer & \multicolumn{5}{c}{\texttt{bert-base-cased}} \\
Transformer Layers & \multicolumn{5}{c}{12} \\
Transformer Dim & \multicolumn{5}{c}{768} \\
Self-Attention Heads & \multicolumn{5}{c}{12} \\
Optimizer & \multicolumn{5}{c}{AdamW} \\
Learning Rate & \multicolumn{5}{c}{$2\cdot10^{-4}$} \\
$\beta_1$, $\beta_2$ & \multicolumn{5}{c}{0.9, 0.98} \\
Warmup steps & \multicolumn{5}{c}{5000} \\
LR scheduler & \multicolumn{5}{c}{Constant} \\
Weight decay & \multicolumn{5}{c}{0.01} \\
Gradient clipping & \multicolumn{5}{c}{1} \\
EMA decay & \multicolumn{5}{c}{0.9999} \\
Batch size & 256 & 256 & 512 & 256 & 256 \\
Training steps & 1M & 225k & 150k & 50k & 150k \\
Max input length & — & 512 & 100 & 50 & 40 \\
Max target length & 80 & 64 & 50 & 50 & 40 \\
Generation steps & 200 & 200 & 100 & 100 & 100 \\
$d$ & 5 & 5 & 7 & 5 & 7 \\
$\delta$, $\sigma_{\text{min}}, \sigma_{\text{max}}$ & \multicolumn{5}{c}{1, 1.5, 200} \\
$\tilde{\delta}$ & 1.1 & 0.1 & 0.1 & 0.1 & 0.1 \\
\bottomrule
\end{tabular}
\end{table}

\section{Dataset Statistics}\label{sec:datasets}

\paragraph{ROCStories}
The ROCStories dataset \citep{rocstories} contains 98,161 five-sentence commonsense fictional stories that capture causal and temporal relations between everyday events. It is a widely used small-scale benchmark for unconditional text generation. The dataset is split into 93,161 training instances, 4,000 validation instances, and 1,000 test instances. 
Url: \texttt{https://cs.rochester.edu/nlp/rocstories/}

\paragraph{XSum}
The XSum dataset \citep{xsum} is used for extreme summarization of BBC news articles. Each article covers a diverse range of topics (e.g., sports, politics) and is paired with a single-sentence summary. The dataset is divided into 204,045 training, 11,332 validation, and 11,334 test instances.    
Url: \texttt{https://huggingface.co/datasets/EdinburghNLP/xsum}

\paragraph{Quasar-T}
Quasar-T \citep{quasar-t} is a large-scale dataset for the question generation task. It requires models to comprehend natural language queries and extract answers from a large corpus. The dataset consists of open-domain trivia questions and their corresponding answers, collected from various internet sources. We use the version preprocessed by \citet{diffuseq}, which includes 116,953 training instances, 2,048 validation instances, and 10,000 test instances.    
Url: \texttt{https://github.com/Shark-NLP/DiffuSeq/tree/main}


\paragraph{QQP}
The Quora Question Pairs (QQP) dataset \citep{qqp} consists of over 400,000 question pairs from the Quora platform, each annotated with a binary label indicating whether the two questions are paraphrases. For the paraphrase generation task, we use the subset containing 149,263 positively labeled pairs, split into 119,410 training instances, 14,926 validation instances, and 14,927 test instances.   
Url: \texttt{https://huggingface.co/datasets/nyu-mll/glue/viewer/qqp}

\paragraph{ParaDetox} 
We use ParaDetox dataset \citep{paradetox} for small-scale conditional generation. It comprises 19,766 pairs of toxic and neutral comments and is intended for the text detoxification task.    
Url: \texttt{https://huggingface.co/datasets/s-nlp/paradetox}

\section{Semantic degradation examples}\label{sec:semantic_degradation}

In Table \ref{tab:degradation_examples}, we provide an example showing how decoded text degrades as the noise level $t$ increases. Note that since the diffusion operates in latent space, these decoded texts are approximate. We obtained them by decoding the model predictions for noisy latents sampled with a forward process at each timestep. Nevertheless, the semantic degradation pattern is clearly visible. At low noise levels, the model first substitutes semantically similar content (numbers, synonyms, related proper nouns), then structural coherence degrades, and finally text collapses into high-frequency tokens.

\begin{table}
    \caption{An example of text semantic degradation obtained with a forward process (Eq. \ref{eq:forward}).}
    \label{tab:degradation_examples}
    \begin{tabularx}{\textwidth}{@{} >{\bfseries}l X @{}}
    \toprule
    $t = 0$ & My friend Jim \textbf{seemed happily married}. He and his wife had \textbf{three sons} and seemedate. They left \textbf{our area} and moved to \textbf{Illinois}. A year later I found out they had gotten divorced. I was shocked and surprised. \\     
    \addlinespace
    $t = 0.6$ & My friend Jim \textbf{married}. He and his wife had \textbf{a son} and feltate. They left \textbf{their city} and moved to \textbf{California}. A year later I found out she had gotten divorced. He was shocked and happy. \\ 
    \addlinespace
    $t = 0.7$ & My friend Jim \textbf{was getting married}. He and his wife used his daughter in my. They stopped the the and wanted to. The year later he found out he had gotten divorced. He was \textbf{excited} and happy. \\
    \addlinespace
    $t = 0.8$ & My friend the the,. He and his wife had a brother of the. She asked the the and wanted. She.. was the the. the. He were happy to him. \\
    \addlinespace
    $t = 0.9$ & John was a the the in the the.. She was the the, the the. He was the the the the. He. the the the.. the. she the the the the. \\
    \bottomrule
    \end{tabularx}
\end{table}

%% file: bibliography.bib
@inproceedings{bert,
    title = "{BERT}: Pre-training of Deep Bidirectional Transformers for Language Understanding",
    author = "Devlin, Jacob  and
      Chang, Ming-Wei  and
      Lee, Kenton  and
      Toutanova, Kristina",
    editor = "Burstein, Jill  and
      Doran, Christy  and
      Solorio, Thamar",
    booktitle = "Proceedings of the 2019 Conference of the North {A}merican Chapter of the Association for Computational Linguistics: Human Language Technologies, Volume 1 (Long and Short Papers)",
    month = jun,
    year = "2019",
    address = "Minneapolis, Minnesota",
    publisher = "Association for Computational Linguistics",
    url = "https://aclanthology.org/N19-1423",
    doi = "10.18653/v1/N19-1423",
    pages = "4171--4186",
}

@misc{stable_audio,
      title={Fast Timing-Conditioned Latent Audio Diffusion}, 
      author={Zach Evans and CJ Carr and Josiah Taylor and Scott H. Hawley and Jordi Pons},
      year={2024},
      eprint={2402.04825},
      archivePrefix={arXiv},
      primaryClass={cs.SD}
}

@misc{stable_video_diffusion,
      title={Stable Video Diffusion: Scaling Latent Video Diffusion Models to Large Datasets}, 
      author={Andreas Blattmann and Tim Dockhorn and Sumith Kulal and Daniel Mendelevitch and Maciej Kilian and Dominik Lorenz and Yam Levi and Zion English and Vikram Voleti and Adam Letts and Varun Jampani and Robin Rombach},
      year={2023},
      eprint={2311.15127},
      archivePrefix={arXiv},
      primaryClass={cs.CV}
}

@inproceedings{latentdiffusion,
  author = {Rombach, Robin and Blattmann, Andreas and Lorenz, Dominik and Esser, Patrick and Ommer, Bj{\"o}rn},
  booktitle = {Proceedings of the IEEE/CVF Conference on Computer Vision and Pattern Recognition},
  pages = {10684--10695},
  title = {High-resolution image synthesis with latent diffusion models},
  year = 2022
}

@misc{sdxl,
      title={SDXL: Improving Latent Diffusion Models for High-Resolution Image Synthesis}, 
      author={Dustin Podell and Zion English and Kyle Lacey and Andreas Blattmann and Tim Dockhorn and Jonas Müller and Joe Penna and Robin Rombach},
      year={2023},
      eprint={2307.01952},
      archivePrefix={arXiv},
      primaryClass={cs.CV},
      url={https://arxiv.org/abs/2307.01952}, 
}

@inproceedings{diffusionbert,
    title = "{D}iffusion{BERT}: Improving Generative Masked Language Models with Diffusion Models",
    author = "He, Zhengfu  and
      Sun, Tianxiang  and
      Tang, Qiong  and
      Wang, Kuanning  and
      Huang, Xuanjing  and
      Qiu, Xipeng",
    editor = "Rogers, Anna  and
      Boyd-Graber, Jordan  and
      Okazaki, Naoaki",
    booktitle = "Proceedings of the 61st Annual Meeting of the Association for Computational Linguistics (Volume 1: Long Papers)",
    month = jul,
    year = "2023",
    address = "Toronto, Canada",
    publisher = "Association for Computational Linguistics",
    url = "https://aclanthology.org/2023.acl-long.248/",
    doi = "10.18653/v1/2023.acl-long.248",
    pages = "4521--4534"
}

@inproceedings{d3pm,
 author = {Austin, Jacob and Johnson, Daniel D. and Ho, Jonathan and Tarlow, Daniel and van den Berg, Rianne},
 booktitle = {Advances in Neural Information Processing Systems},
 editor = {M. Ranzato and A. Beygelzimer and Y. Dauphin and P.S. Liang and J. Wortman Vaughan},
 pages = {17981--17993},
 publisher = {Curran Associates, Inc.},
 title = {Structured Denoising Diffusion Models in Discrete State-Spaces},
 url = {https://proceedings.neurips.cc/paper_files/paper/2021/file/958c530554f78bcd8e97125b70e6973d-Paper.pdf},
 volume = {34},
 year = {2021}
}

@misc{sedd,
      title={Discrete Diffusion Modeling by Estimating the Ratios of the Data Distribution}, 
      author={Aaron Lou and Chenlin Meng and Stefano Ermon},
      year={2024},
      eprint={2310.16834},
      archivePrefix={arXiv},
      primaryClass={stat.ML},
      url={https://arxiv.org/abs/2310.16834}, 
}

@inproceedings{diffusion-lm,
 author = {Li, Xiang and Thickstun, John and Gulrajani, Ishaan and Liang, Percy S and Hashimoto, Tatsunori B},
 booktitle = {Advances in Neural Information Processing Systems},
 editor = {S. Koyejo and S. Mohamed and A. Agarwal and D. Belgrave and K. Cho and A. Oh},
 pages = {4328--4343},
 publisher = {Curran Associates, Inc.},
 title = {Diffusion-LM Improves Controllable Text Generation},
 volume = {35},
 year = {2022}
}

@inproceedings{
ihdm,
title={Generative Modelling with Inverse Heat Dissipation},
author={Severi Rissanen and Markus Heinonen and Arno Solin},
booktitle={The Eleventh International Conference on Learning Representations },
year={2023},
url={https://openreview.net/forum?id=4PJUBT9f2Ol}
}

@inproceedings{tess,
    title = "{TESS}: Text-to-Text Self-Conditioned Simplex Diffusion",
    author = "Karimi Mahabadi, Rabeeh  and
      Ivison, Hamish  and
      Tae, Jaesung  and
      Henderson, James  and
      Beltagy, Iz  and
      Peters, Matthew  and
      Cohan, Arman",
    editor = "Graham, Yvette  and
      Purver, Matthew",
    booktitle = "Proceedings of the 18th Conference of the European Chapter of the Association for Computational Linguistics (Volume 1: Long Papers)",
    month = mar,
    year = "2024",
    address = "St. Julian{'}s, Malta",
    publisher = "Association for Computational Linguistics",
    url = "https://aclanthology.org/2024.eacl-long.144",
    pages = "2347--2361",
}

@inproceedings{han2022ssd,
    title = "{SSD}-{LM}: Semi-autoregressive Simplex-based Diffusion Language Model for Text Generation and Modular Control",
    author = "Han, Xiaochuang  and
      Kumar, Sachin  and
      Tsvetkov, Yulia",
    editor = "Rogers, Anna  and
      Boyd-Graber, Jordan  and
      Okazaki, Naoaki",
    booktitle = "Proceedings of the 61st Annual Meeting of the Association for Computational Linguistics (Volume 1: Long Papers)",
    month = jul,
    year = "2023",
    address = "Toronto, Canada",
    publisher = "Association for Computational Linguistics",
    url = "https://aclanthology.org/2023.acl-long.647",
    doi = "10.18653/v1/2023.acl-long.647",
    pages = "11575--11596",
}

@inproceedings{j-score,
    title = "Reformulating Unsupervised Style Transfer as Paraphrase Generation",
    author = "Krishna, Kalpesh  and
      Wieting, John  and
      Iyyer, Mohit",
    editor = "Webber, Bonnie  and
      Cohn, Trevor  and
      He, Yulan  and
      Liu, Yang",
    booktitle = "Proceedings of the 2020 Conference on Empirical Methods in Natural Language Processing (EMNLP)",
    month = nov,
    year = "2020",
    address = "Online",
    publisher = "Association for Computational Linguistics",
    url = "https://aclanthology.org/2020.emnlp-main.55/",
    doi = "10.18653/v1/2020.emnlp-main.55",
    pages = "737--762"
}

@misc{tencdm,
      title={TEncDM: Understanding the Properties of the Diffusion Model in the Space of Language Model Encodings}, 
      author={Alexander Shabalin and Viacheslav Meshchaninov and Egor Chimbulatov and Vladislav Lapikov and Roman Kim and Grigory Bartosh and Dmitry Molchanov and Sergey Markov and Dmitry Vetrov},
      year={2025},
      eprint={2402.19097},
      archivePrefix={arXiv},
      primaryClass={cs.CL},
      url={https://arxiv.org/abs/2402.19097}, 
}

@inproceedings{diffuseq,
title={DiffuSeq: Sequence to Sequence Text Generation with Diffusion Models},
author={Shansan Gong and Mukai Li and Jiangtao Feng and Zhiyong Wu and Lingpeng Kong},
booktitle={The Eleventh International Conference on Learning Representations },
year={2023},
url={https://openreview.net/forum?id=jQj-_rLVXsj}
}

@inproceedings{gan,
 author = {Goodfellow, Ian J. and Pouget-Abadie, Jean and Mirza, Mehdi and Xu, Bing and Warde-Farley, David and Ozair, Sherjil and Courville, Aaron and Bengio, Yoshua},
 booktitle = {Advances in Neural Information Processing Systems},
 editor = {Z. Ghahramani and M. Welling and C. Cortes and N. Lawrence and K.Q. Weinberger},
 pages = {},
 publisher = {Curran Associates, Inc.},
 title = {Generative Adversarial Nets},
 url = {https://proceedings.neurips.cc/paper_files/paper/2014/file/f033ed80deb0234979a61f95710dbe25-Paper.pdf},
 volume = {27},
 year = {2014}
}

@inproceedings{norm_flow,
author = {Rezende, Danilo Jimenez and Mohamed, Shakir},
title = {Variational inference with normalizing flows},
year = {2015},
publisher = {JMLR.org},
booktitle = {Proceedings of the 32nd International Conference on International Conference on Machine Learning - Volume 37},
pages = {1530–1538},
numpages = {9},
location = {Lille, France},
series = {ICML'15}
}

@article{euclidean_space_hyp,
    title = "Word Embeddings as Metric Recovery in Semantic Spaces",
    author = "Hashimoto, Tatsunori B.  and
      Alvarez-Melis, David  and
      Jaakkola, Tommi S.",
    editor = "Lee, Lillian  and
      Johnson, Mark  and
      Toutanova, Kristina",
    journal = "Transactions of the Association for Computational Linguistics",
    volume = "4",
    year = "2016",
    address = "Cambridge, MA",
    publisher = "MIT Press",
    url = "https://aclanthology.org/Q16-1020/",
    doi = "10.1162/tacl_a_00098",
    pages = "273--286"
}

@inproceedings{multinomial_diffusion,
 author = {Hoogeboom, Emiel and Nielsen, Didrik and Jaini, Priyank and Forr\'{e}, Patrick and Welling, Max},
 booktitle = {Advances in Neural Information Processing Systems},
 editor = {M. Ranzato and A. Beygelzimer and Y. Dauphin and P.S. Liang and J. Wortman Vaughan},
 pages = {12454--12465},
 publisher = {Curran Associates, Inc.},
 title = {Argmax Flows and Multinomial Diffusion: Learning Categorical Distributions},
 url = {https://proceedings.neurips.cc/paper_files/paper/2021/file/67d96d458abdef21792e6d8e590244e7-Paper.pdf},
 volume = {34},
 year = {2021}
}

@inproceedings{diffuseq2,
    title = "{D}iffu{S}eq-v2: Bridging Discrete and Continuous Text Spaces for Accelerated {S}eq2{S}eq Diffusion Models",
    author = "Gong, Shansan  and
      Li, Mukai  and
      Feng, Jiangtao  and
      Wu, Zhiyong  and
      Kong, Lingpeng",
    editor = "Bouamor, Houda  and
      Pino, Juan  and
      Bali, Kalika",
    booktitle = "Findings of the Association for Computational Linguistics: EMNLP 2023",
    month = dec,
    year = "2023",
    address = "Singapore",
    publisher = "Association for Computational Linguistics",
    url = "https://aclanthology.org/2023.findings-emnlp.660/",
    doi = "10.18653/v1/2023.findings-emnlp.660",
    pages = "9868--9875"
}

@inproceedings{
blurring_diffusion,
title={Blurring Diffusion Models},
author={Emiel Hoogeboom and Tim Salimans},
booktitle={The Eleventh International Conference on Learning Representations },
year={2023},
url={https://openreview.net/forum?id=OjDkC57x5sz}
}

@inproceedings{ddpm,
  author = {Ho, Jonathan and Jain, Ajay and Abbeel, Pieter},
  booktitle = {Advances in Neural Information Processing Systems},
  editor = {Larochelle, H. and Ranzato, M. and Hadsell, R. and Balcan, M.F. and Lin, H.},
  pages = {6840--6851},
  publisher = {Curran Associates, Inc.},
  title = {Denoising Diffusion Probabilistic Models},
  volume = 33,
  year = 2020
}

@inproceedings{xsum,
    title = "Don{'}t Give Me the Details, Just the Summary! Topic-Aware Convolutional Neural Networks for Extreme Summarization",
    author = "Narayan, Shashi  and
      Cohen, Shay B.  and
      Lapata, Mirella",
    editor = "Riloff, Ellen  and
      Chiang, David  and
      Hockenmaier, Julia  and
      Tsujii, Jun{'}ichi",
    booktitle = "Proceedings of the 2018 Conference on Empirical Methods in Natural Language Processing",
    month = oct # "-" # nov,
    year = "2018",
    address = "Brussels, Belgium",
    publisher = "Association for Computational Linguistics",
    url = "https://aclanthology.org/D18-1206",
    doi = "10.18653/v1/D18-1206",
    pages = "1797--1807",
}

@inproceedings{qqp,
  title={Quora Question Pairs},
  author={Zihang Chen and Hongbo Zhang and Xiaoji Zhang and Leqi Zhao},
  year={2017},
  url={https://api.semanticscholar.org/CorpusID:233225749}
}

@article{quasar-t,
  title={Quasar: Datasets for Question Answering by Search and Reading},
  author={Dhingra, Bhuwan and Mazaitis, Kathryn and Cohen, William W},
  journal={arXiv preprint arXiv:1707.03904},
  year={2017}
}

@INPROCEEDINGS{div,
  author={Deshpande, Aditya and Aneja, Jyoti and Wang, Liwei and Schwing, Alexander G. and Forsyth, David},
  booktitle={2019 IEEE/CVF Conference on Computer Vision and Pattern Recognition (CVPR)}, 
  title={Fast, Diverse and Accurate Image Captioning Guided by Part-Of-Speech}, 
  year={2019},
  volume={},
  number={},
  pages={10687-10696},
  doi={10.1109/CVPR.2019.01095}}

@misc{bertscore,
      title={BERTScore: Evaluating Text Generation with BERT}, 
      author={Tianyi Zhang and Varsha Kishore and Felix Wu and Kilian Q. Weinberger and Yoav Artzi},
      year={2020},
      eprint={1904.09675},
      archivePrefix={arXiv},
      primaryClass={cs.CL},
      url={https://arxiv.org/abs/1904.09675}, 
}

@inproceedings{rouge,
  title={Rouge: A package for automatic evaluation of summaries},
  author={Lin, Chin-Yew},
  booktitle={Text summarization branches out},
  pages={74--81},
  year={2004}
}

@article{flm,
    title={Flow Map Language Models: One-step Language Modeling via Continuous Denoising},
    author={Chanhyuk Lee and Jaehoon Yoo and Manan Agarwal
            and Sheel Shah and Jerry Huang
            and Aditi Raghunathan and Seunghoon Hong
            and Nicholas M. Boffi and Jinwoo Kim},
    journal={arXiv preprint arXiv:2602.16813},
    year={2026},
}

@misc{openwebtext,
    title={OpenWebText Corpus},
    author={Gokaslan, Aaron and Cohen, Vanya and Pavlick, Ellie and Tellex, Stefanie},
    howpublished={\url{http://Skylion007.github.io/OpenWebTextCorpus}},
    year={2019}
}

@inproceedings{
gidd,
title={Generalized Interpolating Discrete Diffusion},
author={Dimitri von R{\"u}tte and Janis Fluri and Yuhui Ding and Antonio Orvieto and Bernhard Sch{\"o}lkopf and Thomas Hofmann},
booktitle={Forty-second International Conference on Machine Learning},
year={2025},
url={https://openreview.net/forum?id=rvZv7sDPV9}
}

@inproceedings{
  duo2,
  title={The Diffusion Duality, Chapter {II}: ${\textbackslash}Psi$-Samplers and Efficient Curriculum},
  author={Justin Deschenaux and Caglar Gulcehre and Subham Sekhar Sahoo},
  booktitle={The Fourteenth International Conference on Learning Representations},
  year={2026},
  url={https://openreview.net/forum?id=RSIoYWIzaP}
}

@inproceedings{
ddim,
title={Denoising Diffusion Implicit Models},
author={Jiaming Song and Chenlin Meng and Stefano Ermon},
booktitle={International Conference on Learning Representations},
year={2021},
url={https://openreview.net/forum?id=St1giarCHLP}
}

@inproceedings{bleu,
    title = "{B}leu: a Method for Automatic Evaluation of Machine Translation",
    author = "Papineni, Kishore  and
      Roukos, Salim  and
      Ward, Todd  and
      Zhu, Wei-Jing",
    editor = "Isabelle, Pierre  and
      Charniak, Eugene  and
      Lin, Dekang",
    booktitle = "Proceedings of the 40th Annual Meeting of the Association for Computational Linguistics",
    month = jul,
    year = "2002",
    address = "Philadelphia, Pennsylvania, USA",
    publisher = "Association for Computational Linguistics",
    url = "https://aclanthology.org/P02-1040/",
    doi = "10.3115/1073083.1073135",
    pages = "311--318"
}

@article{seqdiffuseq,
  title={SeqDiffuSeq: Text Diffusion with Encoder-Decoder Transformers},
  author={Hongyi Yuan and Zheng Yuan and Chuanqi Tan and Fei Huang and Songfang Huang},
  journal={ArXiv},
  year={2022},
  volume={abs/2212.10325}
}

@inproceedings{genie,
author = {Lin, Zhenghao and Gong, Yeyun and Shen, Yelong and Wu, Tong and Fan, Zhihao and Lin, Chen and Duan, Nan and Chen, Weizhu},
title = {Text generation with diffusion language models: a pre-training approach with continuous paragraph denoise},
year = {2023},
publisher = {JMLR.org},
booktitle = {Proceedings of the 40th International Conference on Machine Learning},
articleno = {867},
numpages = {14},
location = {Honolulu, Hawaii, USA},
series = {ICML'23}
}

@inproceedings{
ar-diffusion,
title={{AR}-Diffusion: Auto-Regressive Diffusion Model for Text Generation},
author={Tong Wu and Zhihao Fan and Xiao Liu and Hai-Tao Zheng and Yeyun Gong and yelong shen and Jian Jiao and Juntao Li and zhongyu wei and Jian Guo and Nan Duan and Weizhu Chen},
booktitle={Thirty-seventh Conference on Neural Information Processing Systems},
year={2023},
url={https://openreview.net/forum?id=0EG6qUQ4xE}
}

@inproceedings{bart,
    title = "{BART}: Denoising Sequence-to-Sequence Pre-training for Natural Language Generation, Translation, and Comprehension",
    author = "Lewis, Mike  and
      Liu, Yinhan  and
      Goyal, Naman  and
      Ghazvininejad, Marjan  and
      Mohamed, Abdelrahman  and
      Levy, Omer  and
      Stoyanov, Veselin  and
      Zettlemoyer, Luke",
    editor = "Jurafsky, Dan  and
      Chai, Joyce  and
      Schluter, Natalie  and
      Tetreault, Joel",
    booktitle = "Proceedings of the 58th Annual Meeting of the Association for Computational Linguistics",
    month = jul,
    year = "2020",
    address = "Online",
    publisher = "Association for Computational Linguistics",
    url = "https://aclanthology.org/2020.acl-main.703",
    doi = "10.18653/v1/2020.acl-main.703",
    pages = "7871--7880",
}

@inproceedings{gpt-2,
  title={Language Models are Unsupervised Multitask Learners},
  author={Alec Radford and Jeff Wu and Rewon Child and David Luan and Dario Amodei and Ilya Sutskever},
  year={2019},
  url={https://api.semanticscholar.org/CorpusID:160025533}
}

@article{vaswani2017attention,
  title={Attention is all you need},
  author={Vaswani, Ashish and Shazeer, Noam and Parmar, Niki and Uszkoreit, Jakob and Jones, Llion and Gomez, Aidan N and Kaiser, {\L}ukasz and Polosukhin, Illia},
  journal={Advances in neural information processing systems},
  volume={30},
  year={2017}
}

@misc{roberta,
      title={RoBERTa: A Robustly Optimized BERT Pretraining Approach}, 
      author={Yinhan Liu and Myle Ott and Naman Goyal and Jingfei Du and Mandar Joshi and Danqi Chen and Omer Levy and Mike Lewis and Luke Zettlemoyer and Veselin Stoyanov},
      year={2019},
      eprint={1907.11692},
      archivePrefix={arXiv},
      primaryClass={id='cs.CL' full_name='Computation and Language' is_active=True alt_name='cmp-lg' in_archive='cs' is_general=False description='Covers natural language processing. Roughly includes material in ACM Subject Class I.2.7. Note that work on artificial languages (programming languages, logics, formal systems) that does not explicitly address natural-language issues broadly construed (natural-language processing, computational linguistics, speech, text retrieval, etc.) is not appropriate for this area.'}
}

@inproceedings{mauve,
 author = {Pillutla, Krishna and Swayamdipta, Swabha and Zellers, Rowan and Thickstun, John and Welleck, Sean and Choi, Yejin and Harchaoui, Zaid},
 booktitle = {Advances in Neural Information Processing Systems},
 editor = {M. Ranzato and A. Beygelzimer and Y. Dauphin and P.S. Liang and J. Wortman Vaughan},
 pages = {4816--4828},
 publisher = {Curran Associates, Inc.},
 title = {MAUVE: Measuring the Gap Between Neural Text and Human Text using Divergence Frontiers},
 url = {https://proceedings.neurips.cc/paper_files/paper/2021/file/260c2432a0eecc28ce03c10dadc078a4-Paper.pdf},
 volume = {34},
 year = {2021}
}

@misc{gpvae,
      title={Diverse Text Generation via Variational Encoder-Decoder Models with Gaussian Process Priors}, 
      author={Wanyu Du and Jianqiao Zhao and Liwei Wang and Yangfeng Ji},
      year={2022},
      eprint={2204.01227},
      archivePrefix={arXiv},
      primaryClass={cs.CL},
      url={https://arxiv.org/abs/2204.01227}, 
}

@article{flan,
  title={Scaling instruction-finetuned language models},
  author={Chung, Hyung Won and Hou, Le and Longpre, Shayne and Zoph, Barret and Tay, Yi and Fedus, William and Li, Yunxuan and Wang, Xuezhi and Dehghani, Mostafa and Brahma, Siddhartha and others},
  journal={Journal of Machine Learning Research},
  volume={25},
  number={70},
  pages={1--53},
  year={2024}
}

@misc{cdcd,
      title={Continuous diffusion for categorical data}, 
      author={Sander Dieleman and Laurent Sartran and Arman Roshannai and Nikolay Savinov and Yaroslav Ganin and Pierre H. Richemond and Arnaud Doucet and Robin Strudel and Chris Dyer and Conor Durkan and Curtis Hawthorne and Rémi Leblond and Will Grathwohl and Jonas Adler},
      year={2022},
      eprint={2211.15089},
      archivePrefix={arXiv},
      primaryClass={cs.CL},
      url={https://arxiv.org/abs/2211.15089}, 
}

@misc{self-cond,
      title={Analog Bits: Generating Discrete Data using Diffusion Models with Self-Conditioning}, 
      author={Ting Chen and Ruixiang Zhang and Geoffrey Hinton},
      year={2023},
      eprint={2208.04202},
      archivePrefix={arXiv},
      primaryClass={cs.CV}
}

@inproceedings{rocstories,
  title={A corpus and cloze evaluation for deeper understanding of commonsense stories},
  author={Mostafazadeh, Nasrin and Chambers, Nathanael and He, Xiaodong and Parikh, Devi and Batra, Dhruv and Vanderwende, Lucy and Kohli, Pushmeet and Allen, James},
  booktitle={Proceedings of the 2016 Conference of the North American Chapter of the Association for Computational Linguistics: Human Language Technologies},
  pages={839--849},
  year={2016}
}

@inproceedings{
ho2021classifierfree,
title={Classifier-Free Diffusion Guidance},
author={Jonathan Ho and Tim Salimans},
booktitle={NeurIPS 2021 Workshop on Deep Generative Models and Downstream Applications},
year={2021},
url={https://openreview.net/forum?id=qw8AKxfYbI}
}

@inproceedings{ld4lg,
 author = {Lovelace, Justin and Kishore, Varsha and Wan, Chao and Shekhtman, Eliot and Weinberger, Kilian Q},
 booktitle = {Advances in Neural Information Processing Systems},
 editor = {A. Oh and T. Naumann and A. Globerson and K. Saenko and M. Hardt and S. Levine},
 pages = {56998--57025},
 publisher = {Curran Associates, Inc.},
 title = {Latent Diffusion for Language Generation},
 volume = {36},
 year = {2023}
}

@inproceedings{
mdlm,
title={Simple and Effective Masked Diffusion Language Models},
author={Subham Sekhar Sahoo and Marianne Arriola and Aaron Gokaslan and Edgar Mariano Marroquin and Alexander M Rush and Yair Schiff and Justin T Chiu and Volodymyr Kuleshov},
booktitle={The Thirty-eighth Annual Conference on Neural Information Processing Systems},
year={2024},
url={https://openreview.net/forum?id=L4uaAR4ArM}
}

@inproceedings{paradetox,
    title = "{P}ara{D}etox: Detoxification with Parallel Data",
    author = "Logacheva, Varvara  and
      Dementieva, Daryna  and
      Ustyantsev, Sergey  and
      Moskovskiy, Daniil  and
      Dale, David  and
      Krotova, Irina  and
      Semenov, Nikita  and
      Panchenko, Alexander",
    editor = "Muresan, Smaranda  and
      Nakov, Preslav  and
      Villavicencio, Aline",
    booktitle = "Proceedings of the 60th Annual Meeting of the Association for Computational Linguistics (Volume 1: Long Papers)",
    month = may,
    year = "2022",
    address = "Dublin, Ireland",
    publisher = "Association for Computational Linguistics",
    url = "https://aclanthology.org/2022.acl-long.469/",
    doi = "10.18653/v1/2022.acl-long.469",
    pages = "6804--6818"
}

@misc{cosmos,
      title={Compressed and Smooth Latent Space for Text Diffusion Modeling}, 
      author={Viacheslav Meshchaninov and Egor Chimbulatov and Alexander Shabalin and Aleksandr Abramov and Dmitry Vetrov},
      year={2025},
      eprint={2506.21170},
      archivePrefix={arXiv},
      primaryClass={cs.CL},
      url={https://arxiv.org/abs/2506.21170}, 
}

@inproceedings{
su2022a,
title={A Contrastive Framework for Neural Text Generation},
author={Yixuan Su and Tian Lan and Yan Wang and Dani Yogatama and Lingpeng Kong and Nigel Collier},
booktitle={Advances in Neural Information Processing Systems},
editor={Alice H. Oh and Alekh Agarwal and Danielle Belgrave and Kyunghyun Cho},
year={2022},
url={https://openreview.net/forum?id=V88BafmH9Pj}
}

@inproceedings{glove,
    title = "{G}lo{V}e: Global Vectors for Word Representation",
    author = "Pennington, Jeffrey  and
      Socher, Richard  and
      Manning, Christopher",
    editor = "Moschitti, Alessandro  and
      Pang, Bo  and
      Daelemans, Walter",
    booktitle = "Proceedings of the 2014 Conference on Empirical Methods in Natural Language Processing ({EMNLP})",
    month = oct,
    year = "2014",
    address = "Doha, Qatar",
    publisher = "Association for Computational Linguistics",
    url = "https://aclanthology.org/D14-1162/",
    doi = "10.3115/v1/D14-1162",
    pages = "1532--1543"
}
